\documentclass{article} 
\usepackage[final]{colm2026_conference}
\usepackage{microtype}
\usepackage{hyperref}
\usepackage{url}
\usepackage{booktabs}
\usepackage{lineno}
\definecolor{darkblue}{rgb}{0, 0, 0.5}
\hypersetup{colorlinks=true, citecolor=darkblue, linkcolor=darkblue, urlcolor=darkblue}

\usepackage{amsmath,amsfonts,bm}

\def\eqref#1{equation~\ref{#1}}

\def\1{\bm{1}}

\DeclareMathAlphabet{\mathsfit}{\encodingdefault}{\sfdefault}{m}{sl}
\SetMathAlphabet{\mathsfit}{bold}{\encodingdefault}{\sfdefault}{bx}{n}

\usepackage{latexsym}
\usepackage[T1]{fontenc}
\usepackage[utf8]{inputenc}
\usepackage{inconsolata}
\usepackage{subfiles}
\usepackage{graphicx}
\usepackage{diagbox}
\usepackage{slashbox}
\usepackage{float}
\usepackage{textcomp}
\usepackage{enumitem}
\usepackage{colortbl}
\usepackage{multirow}
\usepackage{subcaption}
\usepackage{amsmath}
\usepackage{dsfont}
\usepackage{algorithm}
\usepackage{algorithmic}
\usepackage{amssymb}
\usepackage{listings}
\usepackage[capitalise]{cleveref}
\usepackage{xcolor}
\usepackage[para]{footmisc}

\newcommand{\Skip}[1]{}

\newcommand{\dotieconcat}[2]{
  \text{\raisebox{.8ex}{$\smallfrown$}}%
}

\NewDocumentCommand{\Bryan}
{ mO{} }{\textcolor{red}{\textsuperscript{\textit{Bryan}}\textsf{\textbf{\small[#1]}}}}
\NewDocumentCommand{\Sohyun}
{ mO{} }{\textcolor{blue}{\textsuperscript{\textit{Sohyun}}\textsf{\textbf{\small[#1]}}}}
\NewDocumentCommand{\yd}
{ mO{} }{\textcolor{pink}{\textsuperscript{\textit{Da}}\textsf{\textbf{\small[#1]}}}}
\definecolor{my_purple}{HTML}{A02B93}
\definecolor{my_blue}{HTML}{00B0F0}
\definecolor{mygreen}{HTML}{00A64F}
\definecolor{myWhite}{HTML}{FFFFFF}
\definecolor{myGreen}{HTML}{28A745}  
\definecolor{myRed}{HTML}{DC3545} 
\newcommand{\Dataset}{\textbf{DialectGen}}

\title{DialectGen: Benchmarking and Improving Dialect Robustness in Multimodal Generation}
\author{Yu Zhou\textsuperscript{\footnotemark[1]\hspace{3.5pt}\footnotemark[2]},  Sohyun An\textsuperscript{\footnotemark[1]},  Haikang Deng\textsuperscript{\footnotemark[1]},  Da Yin,  Clark Peng, \\
\textbf{Cho-Jui Hsieh, Kai-Wei Chang, Nanyun Peng} \\
University of California, Los Angeles \\
\texttt{\{yuzhou, kwchang, violetpeng\}@cs.ucla.edu}
}

\begin{document}
\ifcolmsubmission
\linenumbers
\fi
\maketitle

\begingroup
  \renewcommand\thefootnote{\fnsymbol{footnote}}
  \footnotetext[1]{Core contributors.}
  \footnotetext[2]{Project lead.}
\endgroup

\begin{abstract}

Contact languages like English exhibit rich regional variations in the form of dialects, which are often used by dialect speakers interacting with generative models. However, can multimodal generative models effectively produce content given dialectal textual input? In this work, we study this question by constructing a new large-scale benchmark spanning six common English dialects. We work with dialect speakers to collect and verify over 4,200 unique prompts and evaluate on 17 image and video generative models. Our automatic and human evaluation results show that current state-of-the-art multimodal generative models exhibit 32.26\% to 48.17\% performance degradation when a single dialect word is used in the prompt. Common mitigation methods such as fine-tuning and prompt rewriting can only improve dialect performance by small margins (< 7\%), while potentially incurring significant performance degradation in Standard American English (SAE). To this end, we design a general encoder-based mitigation strategy for multimodal generative models. Our method teaches the model to recognize new dialect features while preserving SAE performance. Experiments on models such as Stable Diffusion 1.5 show that our method is able to simultaneously raise performance on five dialects to be on par with SAE (+34.4\%), while incurring near-zero cost to SAE performance.
Website, code, and data at: \href{https://dialectgen.github.io}{ {\texttt{dialectgen.github.io}}}.

\end{abstract}
\vspace{-1.5em}
\begin{figure}[h]
\centering
    \includegraphics[width=1.0\linewidth]{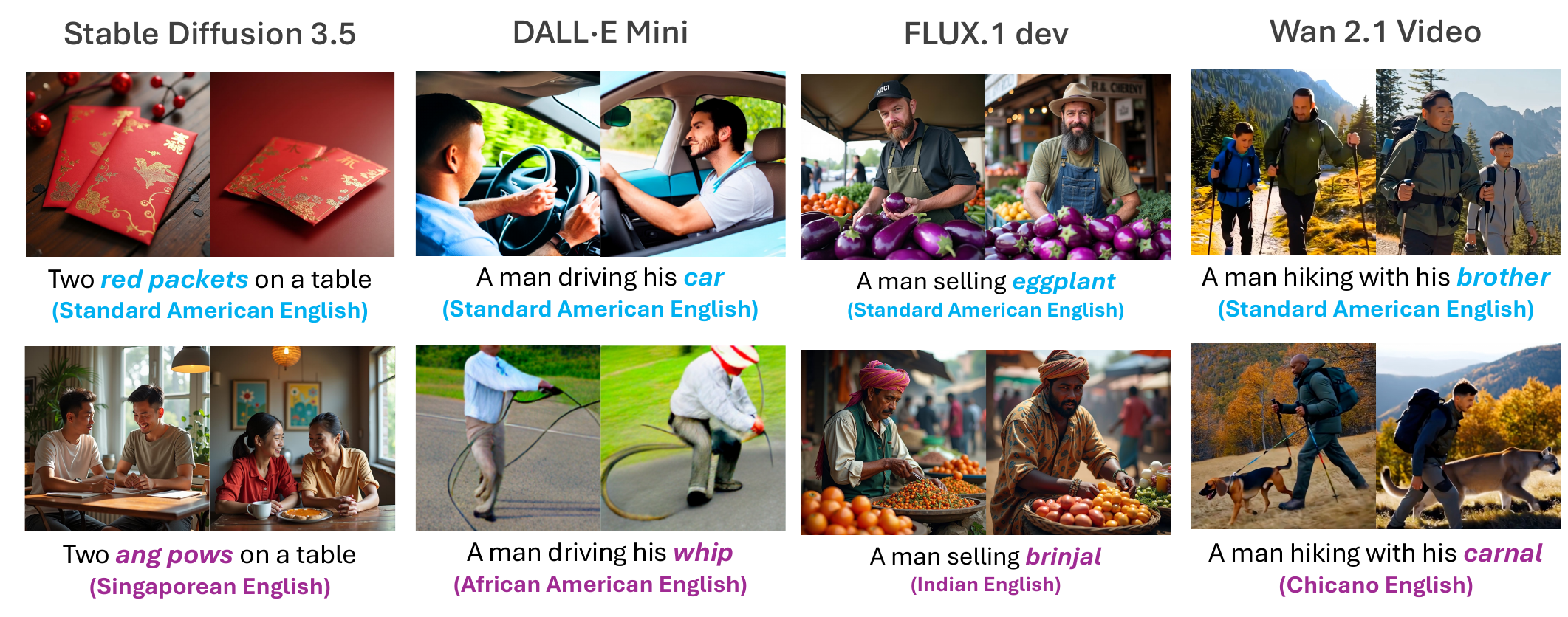}
    \vspace{-1.8em}
    \caption{
    \footnotesize
    \textbf{Model Generations} on semantically identical prompts that differ only in one synonymous lexical feature in \textbf{\textcolor{my_blue}{Standard American English}} (top) / \textbf{\textcolor{my_purple}{a lower-resource English dialect}} (bottom).
    }
    \label{fig:teaser}
    \vspace{-5mm}
\end{figure}
\section{Introduction}
\label{sec:intro}

Linguists have defined over 160 dialects~\citep{aeni2021literature} within the English language, with three out of four English speakers having a dialect background other than Standard American or British English~\citep{crystal2003english}. Despite this rich diversity, current pre-training paradigms employ content filters that can exclude data involving lower-resource English dialects other than Standard American and British English~\citep{gururangan2022whose}, reducing the effectiveness of pretrained models on inputs from other dialects~\citep{lee2023holistic}. Prior works have shown significant allocational harms toward dialect speakers caused by such dialect performance discrepancies in machine learning applications, making the observation of similar performance trends in multimodal generative models an alarming sign~\citep{hovy2016social, bender2021dangers}. 



As shown in \Cref{fig:teaser}, while current multimodal generative models can accurately generate high quality image and video content given Standard American English (SAE) prompts (left), they fail in various manners when provided with semantically equivalent prompts containing a single synonymous dialect word (right). Stable Diffusion 3.5 Large~\citep{esser2024scaling} fails to generate "ang pow", which is commonly used in Singaporean English to mean "red packet", and FLUX.1 [dev]~\citep{flux2024} fails to generate "brinjal", which is synonymous with "eggplant" in Indian English. Furthermore, when the dialect lexeme is polysemous, \textit{i.e.}, has an alternative meaning in SAE, models tend to always generate content that aligns with the SAE meaning, even when the context makes such interpretation highly improbable. For example, DALL-E Mini~\citep{Dayma_DALL·E_Mini_2021} generations of "A man driving his whip" fail to capture the correct meaning of "whip" as "car" in African American English, given clear context indications. 
Similar failure modes are observed in text-to-video generative models: Wan 2.1~\citep{wan2025} fails to correctly render "carnal", which refers to "brother" in Chicano English.



In this work, we construct \Dataset{}, a new large-scale benchmark evaluating dialect robustness in image and video generation. Our benchmark dataset spans six common English dialects, including Standard American English (SAE), African American English (AAE), British English (BrE), Chicano English (ChE), Indian English (InE), and Singaporean English (SgE). For each dialect other than SAE, we create SAE Prompt / Dialect Prompt pairs that are semantically identical besides switching a single SAE lexeme for a synonymous dialect lexeme. We work with dialect speaker annotators to create a rigorous feature selection and prompt filtering pipeline that ensures the final dialect prompts are (1) exactly synonymous with the SAE prompt; (2) valid in the dialect context; and (3) non-ambiguous (for polysemous lexemes). These strictly enforced quality guarantees facilitate the development of simple yet effective automatic and human evaluation metrics for evaluating generative model performance. We experiment with 17 widely used image and video generative models on \Dataset{}, demonstrating up to 38.63\% and 48.17\% performance drops for SOTA open-weights image and video generative models, respectively.



To alleviate such significant dialect performance drops observed in current multimodal generative models, we design a general encoder-based learning strategy that enhances dialect robustness for multimodal generative models. Our method teaches the model's text encoder to recognize dialect lexemes while retaining its knowledge of SAE polysemous lexemes. We also include an encoder-based KL regularization loss based on image-SAE caption datasets to regulate output distribution shifts. Experiments on five dialects show that our method is able to simultaneously improve Stable Diffusion 1.5~\citep{rombach2022high} and SDXL~\citep{podell2023sdxl} performance on five dialects to be on par with SAE performance. At the same time, we observe near zero (< 1\%) SAE performance drop on the general MSCOCO~\citep{lin2014microsoft} validation set for both models.

Our key contributions include:
\begin{itemize}[topsep=0.5em, itemsep=0em, leftmargin=15pt]
    \item \textbf{\Dataset{}}, a new large-scale multi-dialectal benchmark for evaluating dialect robustness in text-to-image and text-to-video generation.
    \item Comprehensive evaluation and analysis of 17 multimodal generative models and five baseline mitigation methods on \Dataset{}.
    \item A high-performing method for improving dialect robustness in multimodal generation while maintaining strong SAE performance.
\end{itemize}

\section{Related Works}
\label{sec:related_works}
\begin{table*}[!t]
\small 
\centering
\begin{tabular}{@{} c c l p{7.1cm} @{}} 
\toprule
\textbf{Dialect} & \textbf{Lexeme} & \textbf{Concise Prompt} & \textbf{Detailed Prompt} \\
\midrule
\textcolor{my_blue}{SAE}
& \textcolor{my_blue}{sneakers} & brand new \textcolor{my_blue}{sneakers} & a little girl wearing a pair of stylish white \textcolor{my_blue}{sneakers} \\
\textcolor{my_purple}{AAE}
& \textcolor{my_purple}{kicks} & brand new \textcolor{my_purple}{kicks} & a little girl wearing a pair of stylish white \textcolor{my_purple}{kicks} \\

\midrule
\textcolor{my_blue}{SAE}
& \textcolor{my_blue}{bathroom} & a spacious \textcolor{my_blue}{bathroom} & a clean and tidy \textcolor{my_blue}{bathroom} with shiny blue wall tiles \\
\textcolor{my_purple}{BrE}
& \textcolor{my_purple}{loo} & a spacious \textcolor{my_purple}{loo} & a clean and tidy \textcolor{my_purple}{loo} with shiny blue wall tiles \\

\midrule
\textcolor{my_blue}{SAE}
& \textcolor{my_blue}{squid} & a \textcolor{my_blue}{squid} on a counter & a large \textcolor{my_blue}{squid} in an aquarium with colorful coral \\
\textcolor{my_purple}{SgE}
& \textcolor{my_purple}{sotong} & a \textcolor{my_purple}{sotong} on a counter & a large \textcolor{my_purple}{sotong} in an aquarium with colorful coral \\

\bottomrule
\end{tabular}
\vspace{-0.3em}
\caption{Example paired textual data entries from \Dataset{}, including \textbf{Lexeme}, \textbf{Concise Prompt}, and \textbf{Detailed Prompt}. Dialect name abbreviations: \textcolor{my_blue}{SAE} (Standard American English), \textcolor{my_purple}{AAE} (African American English), \textcolor{my_purple}{BrE} (British English), \textcolor{my_purple}{SgE} (Singaporean English).}
\label{tab:dataset} 
\end{table*}


Linguists define dialects as regional variations of a language distinguished by unique features in lexicon, phonology, and grammar from each other, together constituting a single language~\citep{hudson1996sociolinguistics, chambers1998dialectology, fromkin1998introduction, nerbonne2009data, wardhaugh2021introduction}. English, like any other language, is subject to such variations. However, most dataset resources and pre-training paradigms focus only on Standard American and British English~\citep{gururangan2022whose}, leading to dialect robustness issues and performance gaps in downstream machine learning applications. Previous works have analyzed and explored such dialectal performance gaps in NLP tasks like QA~\citep{Ziems2023MultiVALUEAF}, NLI~\citep{Ziems2022VALUEUD}, dependency
parsing, and POS tagging~\citep{blodgett2018twitter, jorgensen2015challenges}. Recent works have also noticed the impact of dialect variations on text-to-image generation~\citep{lee2023holistic, wan2024survey}. Along this line of research, we create the first large-scale benchmark of dialect robustness in multimodal generation, evaluating both text-to-image and text-to-video generative models on inputs across six different dialects. Moreover, while lexicon, phonology, and grammar are the three key aspects that distinguish each dialect from others, existing works in Dialectal NLP have so far mainly focused on the grammar variations of dialects~\citep{Ziems2022VALUEUD, Ziems2023MultiVALUEAF, blodgett2018twitter, jorgensen2015challenges}. In this work, we provide the first large-scale dataset of dialectal lexical variations, bridging the gap towards holistic dialectal variation evaluation and building dialect-robust machine learning models.

\section{\Dataset{} Benchmark}
\label{sec:dataset}

\subsection{Dataset Construction}
\label{subsec:Dataset_construction}
To select dialect features for our dataset, we first gather dialect lexemes along with their dictionary definitions and example usages from publicly available dictionaries including The Oxford English Dictionary Regional English Database\footnote{\url{https://www.oed.com/search/advanced/Entries?regional}}, Dictionary of American Regional English~\citep{cassidy1985dictionary}, A Dictionary of Singlish and Singapore English~\citep{lee2004dictionary}, Dictionary of Indian English~\citep{subhash2020dictionary}, and The Oxford Dictionary of African American English~\citep{heinmiller2023compiling}. We collect a total of 1126 dialect lexemes for initial processing.

Based on the dictionary definitions of the selected lexemes, we manually filter out: (1) potentially derogatory lexemes; (2) culture-unique lexemes without Standard American English (SAE) equivalents.
We then carefully read the dictionary definitions of each remaining dialect lexeme and assign it a SAE equivalent lexeme with the same meaning, creating a list of pair-wise corresponding lexical features for each dialect. Examples of selected pairs can be seen in~\Cref{tab:dataset} and~\Cref{fig:teaser}. 

Next, we use GPT4o~\citep{hurst2024gpt4o} to generate prompts for each SAE word in our paired lexical feature set. We specifically instruct the model to generate prompts describing a visual scene with the lexeme playing a central role, which can be one of the following depending on the semantic role of the lexeme: (1) The central object in the scene; (2) The main action of the central object; (3) A prominent descriptive feature of the central object. 

We also ask the model to create two different sets of \textbf{Concise} and \textbf{Detailed} prompts for each SAE lexeme. Then we simply replace the SAE lexeme in the prompts with the dialect lexeme~(\Cref{tab:dataset}) to create our two dialect evaluation settings:
\begin{itemize}[topsep=0.3em, itemsep=0em, leftmargin=10pt]
    \item \textbf{Concise} prompts generally consist of $\leq$ 6 words, with the goal of providing a more challenging evaluation setting where the multimodal generative model is not given too many contextual hints about the lexeme's meaning.
    \item \textbf{Detailed} prompts generally consist of $\geq$ 9 words, with the goal of providing a more relaxed evaluation setting where the multimodal generative model can use more contextual hints to infer the lexeme's meaning.
\end{itemize}

These two evaluation settings also intuitively represent two common user input styles for multimodal generative models, where casual users may tend to provide concise prompts and professional users may be more inclined to write detailed prompts. Across \textbf{Concise} and \textbf{Detailed} evaluation settings, we generate a total of 6552 prompts.



For specific Dialect Prompt / SAE Prompt pairs where the dialect lexeme has an additional polysemous meaning recorded in an SAE dictionary~\citep{webster1869american}, we generate an additional SAE Polysemy Prompt, where the lexeme is used unambiguously in its SAE meaning. This data can be used for regulating model behavior in training scenarios.


\subsection{Dialect Speaker Validation and Filtering}
\label{subsec:Dataset_filtering}

Before admitting the generated prompts to our final evaluation benchmark, we carefully verify their quality and correctness with dialect speaker human annotators. We created a specialized Amazon MTurk interface~(\Cref{fig:mturk}) for prompt annotation and matching potential dialect speaker annotators to their spoken dialect: each human annotator must first self-identify their dialect background and then complete a dialect speaker assessment quiz~\citep{Ziems2023MultiVALUEAF} that matches each annotator to at most one dialect~(\Cref{fig:assessment}). Annotators are only selected if both their self-identified dialect background and their quiz assessment result match to the same dialect. More details on human annotation are available in~\Cref{subsec:human_eval_details}.

After each dialect speaker is selected, they will be presented with Dialect Prompt / SAE Prompt pairs where the only difference is the dialect lexeme being swapped with its SAE equivalent word. For each pair of prompts, the dialect speaker must answer two questions:

\begin{enumerate}[topsep=0.5em, itemsep=0.1em, leftmargin=15pt]
    \item Does the given {Dialect Prompt} make sense in said {Dialect} and correspond exactly in meaning to the given {SAE prompt} in Standard American English? 
    \item Is the given {Dialect Prompt} ambiguous? \textit{i.e.}, Does it have a reasonable alternative interpretation in the Standard American English (SAE) context?
\end{enumerate}

Each Dialect Prompt / SAE Prompt pair is presented to two independent dialect speaker annotators with the choices (Yes / No / I don't know). A pair is included in the final dataset only if both human annotators answer ``Yes'' to the first question and ``No'' to the second question, which ensure the dialect prompt is: (1) exactly synonymous with the SAE prompt. (2) valid in the dialectal context. (3) non-ambiguous (for polysemous lexemes).

In total, dialect speaker filtering further removes 35.9\% of all generated prompts, resulting in a final dataset containing 4,200 validated prompts.

\section{Experiments}
\label{sec:experiments}

\begin{table*}[t!]
\centering
\resizebox{\textwidth}{!}{
\begin{tabular}{ccl|ccc|ccccc}
\toprule
\multicolumn{2}{c}{} & \multirow{2}{*}{\textbf{Model}} & \multicolumn{3}{c|}{\textbf{Overall Performance Drop (\%) $\downarrow$}} & \multicolumn{5}{c}{\textbf{Dialect-wise Performance Drop (\%) $\downarrow$}} \\
\multicolumn{2}{c}{} & & Human & VQAScore & CLIPScore & AAE & BrE & ChE & InE & SgE \\
\midrule
\multirow{18}{*}{\rotatebox[origin=c]{90}{\textbf{Concise Prompts}}}
& \multirow{13}{*}{\rotatebox[origin=c]{90}{\textit{T2I Models}}}

& Stable Diffusion 1.4 & \cellcolor{red!38} 28.19 & \cellcolor{red!36} 26.7 & \cellcolor{red!11} 10.35 & \cellcolor{red!27} 20.67 & \cellcolor{red!10} 9.64 & \cellcolor{red!49} 34.94 & \cellcolor{red!58} 41.27 & \cellcolor{red!37} 26.96 \\
 &  & Stable Diffusion 1.5 & \cellcolor{red!41} 29.77 & \cellcolor{red!37} 27.06 & \cellcolor{red!11} 10.32 & \cellcolor{red!25} 19.51 & \cellcolor{red!9} 8.66 & \cellcolor{red!51} 36.5 & \cellcolor{red!60} 42.15 & \cellcolor{red!39} 28.48 \\
 &  & Stable Diffusion 2.1 & \cellcolor{red!43} 31.46 & \cellcolor{red!39} 28.79 & \cellcolor{red!13} 11.7 & \cellcolor{red!33} 24.35 & \cellcolor{red!10} 9.31 & \cellcolor{red!64} 44.82 & \cellcolor{red!58} 41.12 & \cellcolor{red!39} 28.89 \\
 &  & Stable Diffusion XL & \cellcolor{red!41} 29.8 & \cellcolor{red!36} 26.69 & \cellcolor{red!12} 10.88 & \cellcolor{red!31} 23.37 & \cellcolor{red!8} 7.95 & \cellcolor{red!58} 41.22 & \cellcolor{red!55} 38.74 & \cellcolor{red!29} 22.17 \\
 &  & Stable Diffusion 3 & \cellcolor{red!44} 31.89 & \cellcolor{red!40} 29.01 & \cellcolor{red!12} 10.81 & \cellcolor{red!38} 27.89 & \cellcolor{red!9} 8.64 & \cellcolor{red!60} 42.67 & \cellcolor{red!57} 40.69 & \cellcolor{red!34} 25.12 \\
 &  & Stable Diffusion 3.5 Large & \cellcolor{red!45} 32.31 & \cellcolor{red!40} 29.43 & \cellcolor{red!13} 11.37 & \cellcolor{red!39} 28.3 & \cellcolor{red!10} 9.74 & \cellcolor{red!60} 42.66 & \cellcolor{red!59} 41.9 & \cellcolor{red!33} 24.56 \\
 &  & Stable Diffusion 3.5 Large Turbo & \cellcolor{red!46} 32.92 & \cellcolor{red!42} 30.28 & \cellcolor{red!13} 11.34 & \cellcolor{red!42} 30.33 & \cellcolor{red!10} 9.27 & \cellcolor{red!62} 43.6 & \cellcolor{red!60} 42.49 & \cellcolor{red!35} 25.72 \\
 &  & Flux.1 [dev] & \cellcolor{red!51} 36.43 & \cellcolor{red!45} 32.26 & \cellcolor{red!12} 10.88 & \cellcolor{red!42} 30.61 & \cellcolor{red!12} 10.83 & \cellcolor{red!63} 44.64 & \cellcolor{red!60} 42.59 & \cellcolor{red!45} 32.62 \\
 &  & DALL-E Mini & \cellcolor{red!48} 34.29 & \cellcolor{red!44} 31.52 & \cellcolor{red!13} 11.71 & \cellcolor{red!47} 33.91 & \cellcolor{red!8} 8.18 & \cellcolor{red!67} 47.11 & \cellcolor{red!61} 42.85 & \cellcolor{red!34} 25.51 \\
 &  & DALL-E 2 & \cellcolor{red!54} 38.63 & \cellcolor{red!45} 32.79 & \cellcolor{red!11} 9.97 & \cellcolor{red!50} 35.87 & \cellcolor{red!8} 7.95 & \cellcolor{red!70} 48.78 & \cellcolor{red!67} 47.21 & \cellcolor{red!32} 24.14 \\
 &  & DALL-E 3 & \cellcolor{red!36} 26.55 & \cellcolor{red!33} 24.39 & \cellcolor{red!10} 9.32 & \cellcolor{red!24} 18.97 & \cellcolor{red!1} 3.58 & \cellcolor{red!59} 41.95 & \cellcolor{red!44} 31.9 & \cellcolor{red!34} 25.56 \\
 &  & DALL-E 3 (w/ Prompt Rewrite) & \cellcolor{red!26} 20.19 & \cellcolor{red!23} 18.25 & \cellcolor{red!6} 6.69 & \cellcolor{red!29} 22.11 & \cellcolor{red!5} 6.48 & \cellcolor{red!36} 26.86 & \cellcolor{red!31} 23.05 & \cellcolor{red!15} 12.74 \\
 &  & gpt-image-1 (4o Image Gen)& \cellcolor{red!29} 22.18 & \cellcolor{red!25} 19.18 & \cellcolor{red!7} 7.65 & \cellcolor{red!35} 26.12 & \cellcolor{red!3} 5.2 & \cellcolor{red!35} 26.09 & \cellcolor{red!36} 26.51 & \cellcolor{red!14} 11.99 \\

\cmidrule(lr){2-11}
& \multirow{5}{*}{\rotatebox[origin=c]{90}{\textit{T2V Models}}}

& Cosmos-1 & \cellcolor{red!34} 25.41 & \cellcolor{red!27} 20.49 & \cellcolor{red!6} 6.66 & \cellcolor{red!29} 22.15 & \cellcolor{red!10} 9.69 & \cellcolor{red!35} 26.1 & \cellcolor{red!37} 27.44 & \cellcolor{red!22} 17.09 \\
 &  & Open-Sora & \cellcolor{red!41} 29.98 & \cellcolor{red!36} 26.63 & \cellcolor{red!9} 8.93 & \cellcolor{red!30} 22.59 & \cellcolor{red!9} 9.19 & \cellcolor{red!61} 43.09 & \cellcolor{red!44} 31.74 & \cellcolor{red!36} 26.53 \\
 &  & VideoCrafter-2 & \cellcolor{red!45} 32.5 & \cellcolor{red!42} 30.24 & \cellcolor{red!11} 10.51 & \cellcolor{red!34} 25.36 & \cellcolor{red!10} 9.43 & \cellcolor{red!72} 50.36 & \cellcolor{red!56} 39.95 & \cellcolor{red!35} 26.08 \\
 &  & CogVideoX & \cellcolor{red!57} 40.06 & \cellcolor{red!60} 42.55 & \cellcolor{red!12} 11.04 & \cellcolor{red!54} 38.33 & \cellcolor{red!32} 23.75 & \cellcolor{red!80} 55.18 & \cellcolor{red!35} 54.5 & \cellcolor{red!37} 41.0 \\
 &  & Wan 2.1 & \cellcolor{red!69} 48.17 & \cellcolor{red!68} 47.33 & \cellcolor{red!15} 13.1 & \cellcolor{red!76} 52.68 & \cellcolor{red!43} 31.27 & \cellcolor{red!62} 43.83 & \cellcolor{red!77} 53.38 & \cellcolor{red!80} 55.47 \\

\midrule
\multirow{18}{*}{\rotatebox[origin=c]{90}{\textbf{Detailed Prompts}}}
& \multirow{13}{*}{\rotatebox[origin=c]{90}{\textit{T2I Models}}}

& Stable Diffusion 1.4 & \cellcolor{red!17} 14.33 & \cellcolor{red!20} 15.93 & \cellcolor{red!3} 5.16 & \cellcolor{red!13} 11.65 & \cellcolor{red!2} 4.37 & \cellcolor{red!22} 17.35 & \cellcolor{red!40} 29.23 & \cellcolor{red!21} 17.03 \\
 &  & Stable Diffusion 1.5 & \cellcolor{red!21} 16.56 & \cellcolor{red!20} 16.17 & \cellcolor{red!4} 5.51 & \cellcolor{red!13} 11.18 & \cellcolor{red!4} 5.39 & \cellcolor{red!22} 17.34 & \cellcolor{red!39} 28.7 & \cellcolor{red!23} 18.22 \\
 &  & Stable Diffusion 2.1 & \cellcolor{red!22} 17.39 & \cellcolor{red!24} 18.4 & \cellcolor{red!4} 5.78 & \cellcolor{red!18} 15.06 & \cellcolor{red!4} 5.51 & \cellcolor{red!31} 23.03 & \cellcolor{red!40} 29.36 & \cellcolor{red!25} 19.06 \\
 &  & Stable Diffusion XL & \cellcolor{red!22} 17.12 & \cellcolor{red!22} 17.09 & \cellcolor{red!4} 5.83 & \cellcolor{red!17} 14.09 & \cellcolor{red!4} 5.56 & \cellcolor{red!27} 20.57 & \cellcolor{red!41} 30.12 & \cellcolor{red!18} 15.1 \\
 &  & Stable Diffusion 3 & \cellcolor{red!22} 17.15 & \cellcolor{red!24} 18.64 & \cellcolor{red!4} 5.86 & \cellcolor{red!18} 14.74 & \cellcolor{red!6} 6.67 & \cellcolor{red!32} 23.85 & \cellcolor{red!40} 28.94 & \cellcolor{red!24} 19.02 \\
 &  & Stable Diffusion 3.5 Large & \cellcolor{red!24} 18.42 & \cellcolor{red!25} 19.54 & \cellcolor{red!5} 6.12 & \cellcolor{red!19} 15.7 & \cellcolor{red!6} 6.99 & \cellcolor{red!31} 23.46 & \cellcolor{red!44} 31.83 & \cellcolor{red!26} 19.72 \\
 &  & Stable Diffusion 3.5 Large Turbo & \cellcolor{red!26} 19.9 & \cellcolor{red!27} 20.63 & \cellcolor{red!5} 6.09 & \cellcolor{red!18} 15.06 & \cellcolor{red!8} 8.13 & \cellcolor{red!33} 24.94 & \cellcolor{red!46} 33.42 & \cellcolor{red!28} 21.61 \\
 &  & Flux.1 [dev] & \cellcolor{red!31} 23.29 & \cellcolor{red!28} 21.25 & \cellcolor{red!4} 5.46 & \cellcolor{red!18} 14.84 & \cellcolor{red!9} 9.11 & \cellcolor{red!35} 25.69 & \cellcolor{red!43} 31.4 & \cellcolor{red!34} 25.23 \\
 &  & DALL-E Mini & \cellcolor{red!33} 24.71 & \cellcolor{red!28} 21.44 & \cellcolor{red!6} 7.05 & \cellcolor{red!37} 27.56 & \cellcolor{red!4} 5.29 & \cellcolor{red!37} 27.35 & \cellcolor{red!43} 31.47 & \cellcolor{red!19} 15.53 \\
 &  & DALL-E 2 & \cellcolor{red!22} 17.73 & \cellcolor{red!26} 20.2 & \cellcolor{red!5} 5.98 & \cellcolor{red!24} 18.43 & \cellcolor{red!5} 6.52 & \cellcolor{red!34} 25.5 & \cellcolor{red!45} 32.8 & \cellcolor{red!23} 17.76 \\
 &  & DALL-E 3 & \cellcolor{red!14} 12.18 & \cellcolor{red!16} 13.27 & \cellcolor{red!2} 4.29 & \cellcolor{red!9} 8.85 & \cellcolor{red!3} 4.74 & \cellcolor{red!27} 20.98 & \cellcolor{red!24} 18.91 & \cellcolor{red!15} 12.85 \\
 &  & DALL-E 3 (w/ Prompt Rewrite) & \cellcolor{red!5} 6.55 & \cellcolor{red!12} 10.77 & 2.97 & \cellcolor{red!14} 11.93 & \cellcolor{red!4} 5.28 & \cellcolor{red!12} 10.62 & \cellcolor{red!22} 17.09 & \cellcolor{red!9} 8.94 \\
 &  & gpt-image-1 (4o Image Gen) & \cellcolor{red!9} 8.98 & \cellcolor{red!12} 10.97 & 3.24 & \cellcolor{red!16} 13.72 & \cellcolor{red!2} 4.46 & \cellcolor{red!12} 10.56 & \cellcolor{red!20} 15.96 & \cellcolor{red!11} 10.17 \\

\cmidrule(lr){2-11}
& \multirow{5}{*}{\rotatebox[origin=c]{90}{\textit{T2V Models}}}

& Cosmos-1 & \cellcolor{red!23} 18.04 & \cellcolor{red!17} 14.28 & \cellcolor{red!2} 4.3 & \cellcolor{red!12} 11.05 & \cellcolor{red!10} 9.25 & \cellcolor{red!17} 14.04 & \cellcolor{red!30} 22.49 & \cellcolor{red!18} 14.58 \\
 &  & Open-Sora & \cellcolor{red!22} 17.16 & \cellcolor{red!17} 14.1 & \cellcolor{red!2} 4.57 & \cellcolor{red!16} 13.49 & \cellcolor{red!3} 5.13 & \cellcolor{red!25} 19.4 & \cellcolor{red!26} 19.8 & \cellcolor{red!15} 12.69 \\
 &  & VideoCrafter-2 & \cellcolor{red!30} 22.59 & \cellcolor{red!23} 18.31 & \cellcolor{red!4} 5.91 & \cellcolor{red!21} 16.97 & \cellcolor{red!2} 4.18 & \cellcolor{red!32} 24.16 & \cellcolor{red!38} 27.63 & \cellcolor{red!24} 18.61 \\
 &  & CogVideoX & \cellcolor{red!44} 31.87 & \cellcolor{red!41} 29.6 & \cellcolor{red!8} 8.08 & \cellcolor{red!28} 21.33 & \cellcolor{red!18} 14.63 & \cellcolor{red!45} 32.74 & \cellcolor{red!61} 42.88 & \cellcolor{red!51} 36.4 \\
 &  & Wan 2.1 & \cellcolor{red!45} 32.69 & \cellcolor{red!44} 31.94 & \cellcolor{red!9} 8.59 & \cellcolor{red!42} 30.23 & \cellcolor{red!18} 14.97 & \cellcolor{red!60} 42.58 & \cellcolor{red!51} 36.21 & \cellcolor{red!50} 35.71 \\

\bottomrule
\end{tabular}
}

\caption{\Dataset{} benchmark results for text-to-image and text-to-video generative models, including \textbf{Dialect-wise Performance Drop} measured by VQAScore~\citep{lin2024evaluating}; and \textbf{Overall Performance Drop} measured by human eval, VQAScore, and CLIPScore~\citep{Hessel2021CLIPScoreAR}. Cells are highlighted based on numerical value normalized across the entire table, with darker red indicating a higher performance drop in the given metric.}
\label{tab:main-benchmark-results}
\end{table*}

\subsection{Evaluation Metrics}
\label{subsec:evaluation_metrics}

\paragraph*{Automatic Evaluation}
To automatically evaluate any multimodal generative model $\mathcal{G} (\cdot)$ on our benchmark, we design scoring functions based on reference-free image-text alignment metrics, including VQAScore~\citep{lin2024evaluating} and CLIPScore~\citep{Hessel2021CLIPScoreAR}. For simplicity, we denote any such alignment metric below as $\mathcal{A}$. We further denote the \Dataset{} prompt subset for any dialect as $\mathcal{P}$, which contains many SAE Prompt / Dialect Prompt pairs $p = (p^s,\ p^d)$.

For each individual text prompt $p^s$ or $p^d$, we generate $n$ images under different random seeds for text-to-image generative models, or uniformly sample $n$ frames in a video for text-to-video generative models. Therefore, for each SAE Prompt / Dialect Prompt pair $p = (p^s,\ p^d) \in \mathcal{P}$, we can calculate its SAE and Dialect performance as follows:



\begin{equation}
\mathrm{SAE}(p,\ \mathcal{G}) = \frac{1}{n} \sum_{i=1}^n \mathcal{A}(p^{s},\ \mathcal{G}(p^{s})_i) \qquad \mathrm{Dialect}(p,\ \mathcal{G}) = \frac{1}{n} \sum_{i=1}^n \mathcal{A}(p^{s},\ \mathcal{G}(p^{d})_i)
\label{eq:performance}
\end{equation}

Note that when calculating dialect performance, we align the SAE Prompt $p^s$ with multimodal output generated from the corresponding Dialect Prompt, \textit{i.e.}, $\mathcal{G}(p^d)$. This is feasible given that the paired prompts are synonymous, as verified by dialect speaker annotators in \Cref{subsec:Dataset_filtering}. Based on this, we can compute the dialect-induced performance drop of $\mathcal{G}(\cdot)$ for each prompt pair $p$ before averaging $Drop(p,\ \mathcal{G})$ for all $p$ in $\mathcal{P}$ to obtain $Drop(\mathcal{P},\ \mathcal{G})$.


\begin{equation}
Drop(p,\ \mathcal{G}) = \frac{SAE(p,\ \mathcal{G})-Dialect(p,\ \mathcal{G})}{SAE(p,\ \mathcal{G})}
= \sum_{i=1}^n \frac{\mathcal{A}(\mathcal{G}(p^{s})_i,\ p^{s}) - \mathcal{A}(\mathcal{G}(p^{d})_i,\ p^{s})}{\mathcal{A}(\mathcal{G}(p^{s})_i,\ p^{s})}
\label{eq:drop}
\end{equation}


\vspace{+0.0001em}

\paragraph*{Human Evaluation}


We further design a human evaluation pipeline to check the empirical alignment between our automatic evaluation metrics and human judgment. For 5\% of the model outputs in our benchmark, we ask three independent external human annotators to evaluate: to what extent do the multimodal generations conditioned on the SAE Prompt $\mathcal{G}(p^s)$ or Dialect Prompt $\mathcal{G}(p^d)$ match the scene described by SAE prompt $p^s$. Annotators are asked to rate the alignment between each (image/video, caption) pair with a numerical score between 0 and 10. 
The numerical scores are scaled by 0.1 to match the scoring range of VQAScore and CLIPScore before calculating SAE and Dialect performance. 
Finally, we use the same formula to calculate the dialect-induced performance drop $Drop(p,\ \mathcal{G})$. Since we only evaluate the alignment between image/video and the SAE prompt, this task does not require dialect speaker human annotators.

\subsection{Benchmark Experiments}
\label{subsec:benchmark_experiments}

Using metrics described in \Cref{subsec:evaluation_metrics}, we evaluate popular open-weights and proprietary multimodal generative models on \Dataset{}. Model performances are separately aggregated for \textbf{Concise Prompts} and \textbf{Detailed Prompts} settings in \Cref{tab:main-benchmark-results}.

\paragraph*{Overall Performances}

For each model, we record overall dialect-induced performance drop on \Dataset{} using three different metrics: Human Eval, VQAScore, and CLIPScore. We calculate Pearson correlation coefficients~\citep{pearson1895} $r$ between each of the two metrics and observe $r$(Human, VQAScore) = 0.968, $r$(Human, CLIPScore) = 0.924, and $r$(VQAScore, CLIPScore) = 0.907. This shows that while both automatic scoring metrics have high correlations to human judgement (the gold standard), VQAScore is a better-aligned scoring metric for measuring dialect-induced performance drop.

Contrasting the model performance drops across the two evaluation settings \textbf{Concise Prompts} and \textbf{Detailed Prompts}, we can clearly see that all models exhibit significantly larger performance drops for concise prompts compared to detailed prompts. This is in line with our assumption that models can more easily infer the meanings of unknown dialect lexemes from richer prompt contexts, highlighting the need for challenging evaluation via concise prompts to reveal model robustness issues.

Among text-to-video generative models: Wan 2.1~\citep{wan2025} and CogVideoX~\citep{yang2024cogvideox}
exhibit the largest overall performance drops while Cosmos-1~\citep{agarwal2025cosmos} is the most robust. While for text-to-image generative models, DALL-E 2~\citep{ramesh2022dalle2} and Flux.1 [dev]~\citep{flux2024} exhibit the largest overall performance drops while DALL-E 3~\citep{betker2023improving} (w/ Prompt Rewrite) and gpt-image-1 (4o Image Generation)~\citep{openai2025_4o_image_generation} are the most robust.


\paragraph*{Dialect-wise Performance Drop}

In addition to overall performance, we record each model's performance drops for every dialect. From the color heatmap in \Cref{tab:main-benchmark-results}, we can clearly see that the majority of models experience the most severe performance degradations with ChE and InE input, while also suffering significant performance drops with AAE and SgE input. On the other hand, model performances generally do not degrade significantly for BrE, which can be attributed to BrE being a relatively higher-resource dialect in training datasets. In \Cref{app:text_encoder_data_analysis}, we further analyze lexeme appearance frequencies in training datasets and show correlation with observed performance drops.

\section{Mitigation Methods}
\label{sec:mitigation}
\begin{figure*}[t!]
\centering
    \includegraphics[width=1.0\linewidth]{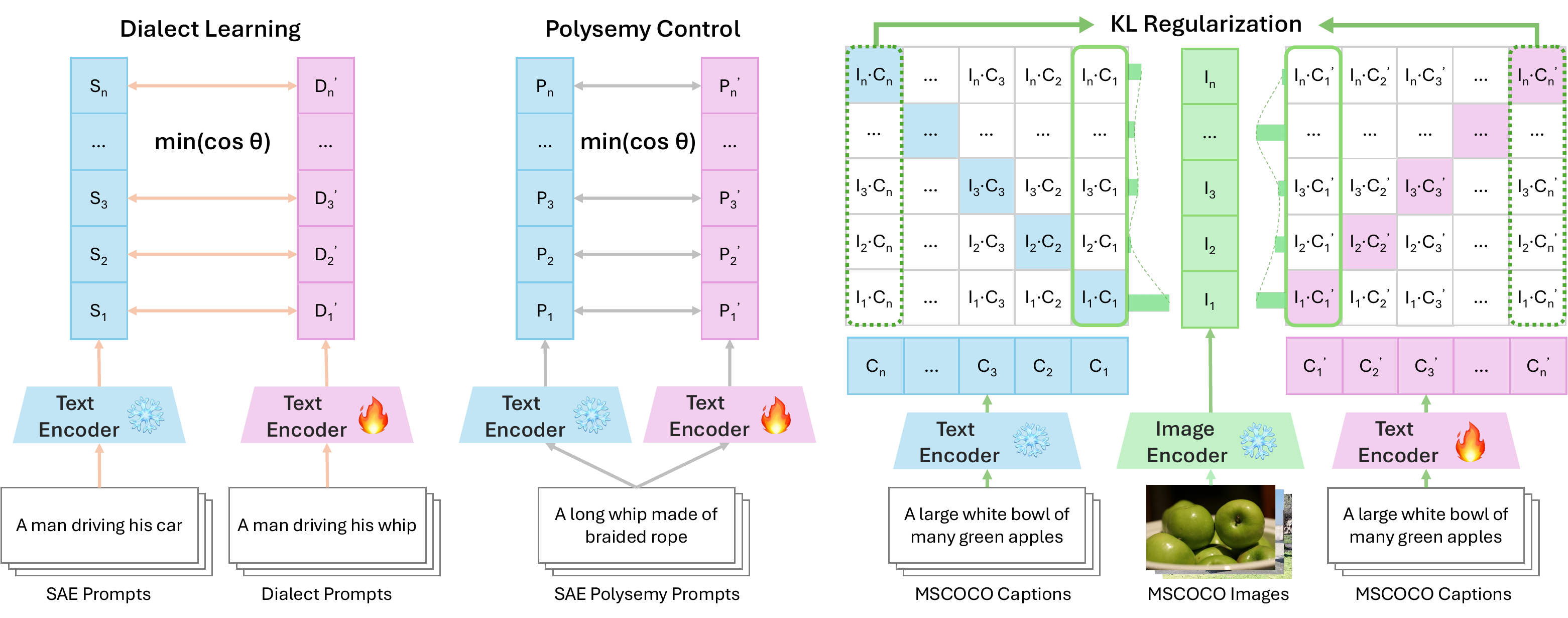}
    \caption{
    Losses used in our mitigation. Text prompts for \textbf{Dialect Learning} and \textbf{Polysemy Control} come from the \Dataset{} training set, while image-caption pairs for \textbf{KL Regularization} come from the MSCOCO validation set.
    }
    \label{fig:method}
\end{figure*}

The significant dialect performance drops of current multimodal generative models shown in \Cref{subsec:benchmark_experiments} highlight the need for effective mitigation strategies to improve dialect robustness. Here, the goal is to develop a method that enhances robustness across multiple dialects while preserving performance on standard SAE prompts. 

To this end, we first investigate intuitive baseline approaches, including (1) Decoder Finetuning; (2) Prompt Revision, and then introduce our new mitigation strategy.

\vspace{-5pt}
\subsection{Baseline Methods}
\label{subsec:baseline_methods}
\paragraph{Decoder Finetuning}
The majority of current text-to-image and text-to-video generative models comprise two main components: a text encoder and an image/video decoder. In current post-training paradigms, typically the text encoder is kept frozen while the decoder is fine-tuned~\citep{podell2023sdxl, rombach2022high, betker2023improving, dai2023emu}. Existing works in aligning, enhancing, and customizing multimodal generative models also focus heavily on developing reward-based fine-tuning methods for the decoder while freezing the text encoder~\citep{Segalis2023API, clark2023directly, prabhudesai2023aligning, black2023training, fan2023dpok, wallace2024diffusion, dang2025personalized}. 

Based on existing works, we apply prominent multimodal generation enhancement methods towards improving dialect robustness, including:
\begin{itemize}[topsep=0.3em, itemsep=0em, leftmargin=10pt]
    \item \textbf{Decoder Fine-tune}~\citep{rombach2022high} given a pair of synonymous Dialect / SAE Prompts, we fine-tune the decoder with the Dialect Prompt as input, and images generated using the SAE Prompt as target output.
    \item \textbf{Diffusion DPO}~\citep{wallace2024diffusion} We use the Dialect Prompt as input, and use images generated with the SAE Prompt / Dialect Prompt as Win / Lose pairs for DPO.
\end{itemize}

\vspace{-5pt}
\paragraph{Prompt Revision}
Beyond decoder fine-tuning, another family of methods for aligning and enhancing multimodal generative models is prompt revision~\citep{hao2023optimizing, betker2023improving, wang2024promptcharm, chen2024tailored}. We experiment with both a general prompt rewriting method and targeted prompt translation methods using general-purpose LLMs:

\begin{itemize}[topsep=0.3em, itemsep=0em, leftmargin=10pt]
    \item \textbf{Prompt Rewrite} We apply the general prompt rewriting pipeline in~\citet{betker2023improving} to all test prompts before passing them to the generative model.
    \item \textbf{Prompt Translate} We use general-purpose LLMs~\citep{grattafiori2024llama, singh2025openai} to translate all prompts to SAE before passing them to the generative model.
\end{itemize}

\vspace{-5pt}
\subsection{Our Method}
\label{subsec:our_methods}
Unlike prior approaches, we propose a new mitigation strategy that focuses on updating the text encoder(s). A natural first step toward improving dialectal robustness is to align the semantic representation of a dialect expression with that of its corresponding SAE counterpart.

\paragraph{Dialect Learning} To operationalize this idea, we introduce a Dialect Learning loss that encourages the target text encoder to recognize dialectal lexemes by minimizing the cosine distance between the target encoder’s embedding of a dialect prompt and the frozen encoder’s embedding of its synonymous SAE prompt:
\vspace{-0.5em}
\begin{equation}
\begin{aligned}  
    \mathcal{L}_{\text{DL}} = \frac{1}{N} \sum_{i=1}^{N} \left( 1 - \langle \pi(p^d_i),\ \pi_0(p^s_i) \rangle \right).
    \label{eq:loss-dialect}
\end{aligned}
\end{equation}
Here, $\langle \cdot,\ \cdot \rangle$ denotes cosine similarity; $\pi(\cdot)$ and $\pi_0(\cdot)$ represent the trainable target text encoder and the frozen reference encoder, respectively; and $p^d_i$ and $p^s_i$ denote the $i$-th pair of synonymous prompts in dialect and standard English, respectively.
Although this may improve dialectal robustness, relying on this loss alone may compromise the model’s ability to handle dialect lexemes that exhibit polysemy in SAE contexts.

\paragraph{Polysemy Control} In order to retain the model's ability to correctly recognize polysemous lexemes within SAE contexts, we introduce a Polysemy Control loss that minimizes the cosine distance between embeddings of the same SAE polysemous prompt generated by the target and frozen encoders:
\vspace{-0.3em}
\begin{equation}
\begin{aligned}
    \mathcal{L}_{\text{PC}} = \frac{1}{N} \sum_{i=1}^{N} \left( 1 - \langle \pi(p^m_i),\ \pi_0(p^m_i) \rangle \right),
    \label{eq:loss-polysemy}
\end{aligned}
\end{equation}
where each $p^m_i$ is a polysemous SAE prompt sampled from the dataset. This loss is applied only to examples containing SAE polysemous lexemes.
\vspace{-5pt}
\begin{table*}[t]

\centering
\small
\resizebox{\textwidth}{!}{
\begin{tabular}{l|ccc|cccccc}
\toprule
\multirow{3}{*}{\textbf{Mitigation Methods}} & \multicolumn{3}{c|}{\textbf{Overall Performances $\uparrow$}} & \multicolumn{5}{c}{\textbf{Dialect Performance} $\uparrow$} \\
& SAE & SAE & Dialect & \multirow{2}{*}{AAE} & \multirow{2}{*}{BrE} & \multirow{2}{*}{ChE} & \multirow{2}{*}{InE} & \multirow{2}{*}{SgE} \\
& MSCOCO & Polysemy & Avg. & & & & & \\
\midrule
\textbf{Base Model (Stable Diffusion 1.5)} & \cellcolor{mygreen!50} 75.49 & \cellcolor{mygreen!50} 72.84 & 57.8 & \cellcolor{mygreen!7} 60.13 & \cellcolor{mygreen!3} 69.39 & 52.65 & 49.94 & 56.89 \\
\midrule
\textbf{Prompt Revision} &  &  &  &  &  &  &  &  \\
\hspace{1em}DALL-E 3 Prompt Rewrite & \cellcolor{mygreen!44} 74.03 & \cellcolor{mygreen!47} 71.33 & \cellcolor{mygreen!2} 58.48 & 57.73 & \cellcolor{mygreen!8} 70.4 & \cellcolor{mygreen!2} 53.98 & \cellcolor{mygreen!1} 50.42 & \cellcolor{mygreen!6} 59.87 \\
\hspace{1em}LLaMA 3 Prompt Translate & \cellcolor{mygreen!45} 74.25 & \cellcolor{mygreen!46} 70.85 & \cellcolor{mygreen!8} 60.91 & \cellcolor{mygreen!9} 60.87 & \cellcolor{mygreen!29} 74.39 & \cellcolor{mygreen!12} 59.05 & \cellcolor{mygreen!5} 53.04 & \cellcolor{mygreen!17} 64.98 \\
\hspace{1em}GPT 5 Prompt Translate & \cellcolor{mygreen!45} 74.3 & \cellcolor{mygreen!47} 71.08 & \cellcolor{mygreen!17} 64.6 & \cellcolor{mygreen!13} 62.35 & \cellcolor{mygreen!23} 73.18 & \cellcolor{mygreen!14} 59.91 & \cellcolor{mygreen!17} 60.22 & \cellcolor{mygreen!21} 67.33 \\
\textbf{Decoder Fine-tuning} &  &  &  &  &  &  &  &  \\
\hspace{1em}Decoder Finetune & \cellcolor{mygreen!5} 65.01 & \cellcolor{mygreen!11} 52.13 & \cellcolor{mygreen!8} 60.94 & \cellcolor{mygreen!17} 63.85 & \cellcolor{mygreen!7} 70.14 & \cellcolor{mygreen!9} 57.3 & \cellcolor{mygreen!5} 52.84 & \cellcolor{mygreen!8} 60.56 \\
\hspace{1em}Diffusion DPO & 63.94 & \cellcolor{mygreen!8} 50.32 & \cellcolor{mygreen!14} 63.52 & \cellcolor{mygreen!24} 66.31 & 68.91 & \cellcolor{mygreen!16} 61.22 & \cellcolor{mygreen!10} 56.38 & \cellcolor{mygreen!16} 64.79 \\
\textbf{Our Encoder Tuning Methods} &  &  &  &  &  &  &  &  \\
\hspace{1em}Dialect Learning & \cellcolor{mygreen!14} 67.14 & 46.3 & \cellcolor{mygreen!50} 78.02 & \cellcolor{mygreen!49} 75.21 & \cellcolor{mygreen!50} 78.33 & \cellcolor{mygreen!50} 79.31 & \cellcolor{mygreen!46} 78.1 & \cellcolor{mygreen!46} 79.15 \\
\hspace{1.5em}+ Text Cosine Reg. & \cellcolor{mygreen!14} 67.06 & 46.39 & \cellcolor{mygreen!50} 77.93 & \cellcolor{mygreen!50} 75.44 & \cellcolor{mygreen!47} 77.84 & \cellcolor{mygreen!50} 79.31 & \cellcolor{mygreen!46} 78.22 & \cellcolor{mygreen!45} 78.86 \\
\hspace{1.5em}+ Image Cosine Reg. & \cellcolor{mygreen!16} 67.73 & 46.48 & \cellcolor{mygreen!50} 78 & \cellcolor{mygreen!49} 74.91 & \cellcolor{mygreen!49} 78.2 & \cellcolor{mygreen!50} 79.45 & \cellcolor{mygreen!46} 78.33 & \cellcolor{mygreen!46} 79.11 \\
\hspace{1.5em}+ Text KL Reg. & \cellcolor{mygreen!38} 72.68 & \cellcolor{mygreen!12} 52.72 & \cellcolor{mygreen!49} 77.78 & \cellcolor{mygreen!47} 74.4 & \cellcolor{mygreen!50} 78.27 & \cellcolor{mygreen!48} 78.36 & \cellcolor{mygreen!46} 78.17 & \cellcolor{mygreen!47} 79.71 \\
\hspace{1.5em}+ Image KL Reg. & \cellcolor{mygreen!34} 71.69 & \cellcolor{mygreen!13} 53.41 & \cellcolor{mygreen!50} 78.12 & \cellcolor{mygreen!45} 73.77 & \cellcolor{mygreen!44} 77.23 & \cellcolor{mygreen!49} 79.06 & \cellcolor{mygreen!48} 79.25 & \cellcolor{mygreen!50} 81.29 \\
\hspace{1.5em}+ Text KL Reg. + Polysemy Ctrl. & \cellcolor{mygreen!38} 72.71 & \cellcolor{mygreen!45} 70.15 & \cellcolor{mygreen!49} 77.74 & \cellcolor{mygreen!41} 72.24 & \cellcolor{mygreen!36} 75.76 & \cellcolor{mygreen!49} 78.95 & \cellcolor{mygreen!50} 80.67 & \cellcolor{mygreen!50} 81.07 \\
\hspace{1.5em}+ Image KL Reg. + Polysemy Ctrl. & \cellcolor{mygreen!47} 74.8 & \cellcolor{mygreen!47} 71.17 & \cellcolor{mygreen!49} 77.68 & \cellcolor{mygreen!42} 72.61 & \cellcolor{mygreen!42} 76.74 & \cellcolor{mygreen!46} 77.51 & \cellcolor{mygreen!50} 80.41 & \cellcolor{mygreen!50} 81.14 \\

\bottomrule
\end{tabular}
}
\caption{
\textbf{Mitigation results} for all baseline methods and our best performing method, including \textbf{Overall Performances} on SAE MSCOCO, SAE Polysemy, average Dialect performance, and \textbf{Dialect Performance} for each dialect, all measured using VQAScore~\cite{lin2024evaluating}. Cell colors reflect column-normalized performance values, with darker green indicating higher VQAScore performance.
}

\vspace{-0.7em}
\label{tab:mitigation-methods}
\end{table*}

\paragraph{KL Regularization}
In addition to the previous two losses, it is also essential to preserve the model’s performance on general SAE prompts. 
To this end, one might consider employing the conventional Kullback-Leibler (KL) divergence loss, which promotes alignment between the output distributions of a trainable target model and a frozen reference model over a predefined discrete logit space. However, this approach is not directly applicable in our setting, as text encoders output continuous embeddings rather than discrete logits. To address this challenge, we approximate the output distribution by computing similarity scores between a given caption embedding and a set of reference image embeddings drawn from a joint image-text embedding space.
Concretely, we begin by sampling $M$ caption-image pairs $\left\{(x_i^{\text{cap}},\ x_i^{\text{img}}) \mid i \in [M] \right\}$ from a general SAE dataset such as MSCOCO~\citep{lin2014microsoft}. For each pair, we compute the caption embedding $C_i = \pi_0(x_i^{\text{cap}})$ using a frozen text encoder $\pi_0$, and the corresponding image embedding $I_i = \phi_0(x_i^{\text{img}})$ using a frozen image encoder $\phi_0$, with both encoders operating in the same shared text-image embedding space. The resulting image embeddings $\left\{ I_i \mid i \in [M] \right\}$ serve as reference anchors for computing similarity scores with a given caption embedding. These scores act as surrogate logits that approximate the output distributions required for the KL divergence computation. Specifically, for each caption $x_i^{\text{cap}}$, we define the approximated output distributions for the frozen encoder $\pi_0$ and the trainable target encoder $\pi$ as:
\vspace{-0.3em}
\begin{equation}
\mathbf{s}^{\pi_0}_i = \left[ \langle I_1, C_i \rangle,\ \dots,\ \langle I_M, C_i \rangle \right], \qquad \mathbf{s}^{\pi}_i = \left[ \langle I_1, C'_i \rangle,\ \dots,\ \langle I_M, C'_i \rangle \right]
\end{equation}
where $C'_i = \pi(x_i^{\text{cap}})$.

Given these simulated logits, we define the KL divergence loss to encourage the target encoder’s output distribution to remain close to that of the frozen encoder:
\vspace{-0.3em}
\begin{equation}
\begin{aligned}
    \mathcal{L}_{\text{KL}} &= \frac{1}{M} \sum_{i=1}^{M} \text{KL} \left( \operatorname{softmax}(\mathbf{s}^{\pi}_i) \,\|\, \operatorname{softmax}(\mathbf{s}^{\pi_0}_i) \right).
    \label{eq:loss-kl}
\end{aligned}
\end{equation}
This approach is compatible with CLIP-style models~\citep{Radford2021LearningTV, zhai2023sigmoid}, in which image and text embeddings are aligned within a shared representation space. When an image encoder is unavailable, we instead use the frozen caption embeddings $\left\{ C_i \mid i \in [M] \right\}$ as proxies for reference anchors. We hereafter refer to the case where image embeddings are used as reference anchors as ``Image KL Reg.'' and the one using text embeddings as ``Text KL Reg.''

\subsection{Mitigation Results}
\label{subsec:mitigation_experiments}
Here, we validate all baselines and our method on SD1.5, SDXL, and Wan 2.1 Video. Due to space limitations, experiment results for SDXL and Wan 2.1 Video are reported in \Cref{subsec:additional_mitigation-results}.

As shown in \Cref{tab:mitigation-methods}, prompt rewriting methods that operate solely at the input level do not degrade SAE MSCOCO or polysemy performance, but yield only slight improvements up to 6.8\% in average dialect performance. Furthermore, decoder fine-tuning approaches also lead to small gains of up to 5.7\% in dialect performance, but at the cost of substantial drops in both general SAE and polysemy scores. In contrast, our method, corresponding to the last row of the table and incorporating all three loss components described in \Cref{subsec:our_methods}, significantly improves dialect robustness across all five dialects. Its average dialect performance of 77.68\% closely approaches the base model’s SAE average score of 77.91\%, while causing negligible degradation in SAE MSCOCO and polysemy performance. 


\vspace{-7pt}
\paragraph{Base Model vs. Dialect Learning}
As shown in \Cref{tab:mitigation-methods}, applying the Dialect Learning loss ($\mathcal{L}_{\text{DL}}$) alone yields huge improvements in the base model’s dialect performance, but also degrades SAE MSCOCO and polysemy performance.

\vspace{-7pt}
\paragraph{Cosine Reg. vs. KL Reg.}
Simply maximizing cosine similarity between the target text encoder’s text embeddings and the corresponding text/image embeddings from the frozen text/image encoder (denoted as Text/Image Cosine Reg.), which is computed over the same caption-image pairs used in our KL regularization, does not effectively recover the base model’s SAE MSCOCO and polysemy performance. In contrast, adding our KL regularization loss ($\mathcal{L}_{\text{KL}}$) improves both metrics while preserving dialect gains. 

\vspace{-7pt}
\paragraph{Adding Polysemy Ctrl.}
Finally, incorporating the Polysemy Control loss ($\mathcal{L}_{\text{PC}}$) yields substantial gains in polysemy performance, improving it by 17.43\% and 17.76\% for Text and Image KL Reg. respectively, underscoring the importance of this component in recognizing polysemous lexemes within SAE contexts.
\section{Conclusion}
\label{sec:conclusion}
\vspace{-5pt}

In this work, we create \Dataset{}, a large-scale multi-dialectal benchmark evaluating the dialect robustness of multimodal generative models. Our experiments on 17 widely used text-to-image and text-to-video generative models reveal severe performance drops up to 38.63\% and 48.17\% for image and video generative models, respectively. We further design an encoder-based mitigation strategy to enhance dialect robustness while preserving performance on Standard American English.
\section{Acknowledgements}
\label{sec:acknowledgements}

We would like to thank Connor Couture and Allen Cheung for designing the initial version of our MTurk annotation interface, and Xinrong Du for providing feedback. We also thank Prof. Diyi Yang, Prof. Xiang Chen, Caleb Ziems, Julia Kruk, Hritik Bansal, Jiachen Gu, Zongyu Lin, and Amita Kamath for valuable discussions and pointers; and especially Caleb Ziems for sharing the grammar-based dialect-speaker quiz he created. Finally, we appreciate Wenbo Hu, Lucas Bandarkar, Mohsen Fayyaz, and Tanmay Parekh for their helpful feedback on the paper draft. 
\newpage
\bibliography{custom}

@string{ECCV="European Conference on Computer Vision (ECCV)"}

@string{ICML="International Conference on Machine Learning (ICML)"}

@string{ACL="Association for Computational Linguistics (ACL)"}

@article{radford2021learning,
  title={Learning transferable visual models from natural language supervision},
  author={Radford, Alec and Kim, Jong Wook and Hallacy, Chris and Ramesh, Aditya and Goh, Gabriel and Agarwal, Sandhini and Sastry, Girish and Askell, Amanda and Mishkin, Pamela and Clark, Jack and others},
  journal={arXiv preprint arXiv:2103.00020},
  year={2021}
}

@inproceedings{zhai2023sigmoid,
  title={Sigmoid loss for language image pre-training},
  author={Zhai, Xiaohua and Mustafa, Basil and Kolesnikov, Alexander and Beyer, Lucas},
  booktitle={Proceedings of the IEEE/CVF international conference on computer vision},
  pages={11975--11986},
  year={2023}
}

@inproceedings{lin2014microsoft,
  title={Microsoft coco: Common objects in context},
  author={Lin, Tsung-Yi and Maire, Michael and Belongie, Serge and Hays, James and Perona, Pietro and Ramanan, Deva and Doll{\'a}r, Piotr and Zitnick, C Lawrence},
  booktitle=ECCV,
  year={2014},
}

@book{wardhaugh2021introduction,
  title={An introduction to sociolinguistics},
  author={Wardhaugh, Ronald and Fuller, Janet M},
  year={2021},
  publisher={John Wiley \& Sons}
}

@book{chambers1998dialectology,
  title={Dialectology},
  author={Chambers, Jack K and Trudgill, Peter},
  year={1998},
  publisher={Cambridge University Press}
}

@book{hudson1996sociolinguistics,
  title={Sociolinguistics},
  author={Hudson, Richard A},
  year={1996},
  publisher={Cambridge university press}
}

@article{nerbonne2009data,
  title={Data-driven dialectology},
  author={Nerbonne, John},
  journal={Language and linguistics compass},
  volume={3},
  number={1},
  pages={175--198},
  year={2009},
  publisher={Wiley Online Library}
}

@article{fromkin1998introduction,
  title={An Introduction to Language 6e},
  author={Fromkin, VA and Rodman, Robert and Hyams, V},
  journal={Orlando, FL: Hartcourt Brace College Publishers},
  year={1998}
}

@book{webster1869american,
  title={An American dictionary of the English language},
  author={Webster, Noah},
  year={1869},
  publisher={Merriam}
}

@book{cassidy1985dictionary,
  title={Dictionary of American Regional English: IO},
  author={Cassidy, Frederic G},
  year={1985},
  publisher={Belknap Press of Harvard University Press}
}

@book{lee2004dictionary,           
  author    = {Lee, Jack Tsen-Ta},
  title     = {A Dictionary of Singlish and Singapore English},
  year      = {2004},
  publisher = {Lee, Jack Tsen-Ta},
  note      = {Accessed 2025-05-16}
}

@book{subhash2020dictionary, 
  author    = {Subhash, V.},
  title     = {Dictionary of Indian English},
  year      = {2020},
  publisher = {V.~Subhash},
  address   = {Hyderabad},
  isbn      = {9789354374487}
}

@article{heinmiller2023compiling,
  title={Compiling The Oxford Dictionary of African American English: A Progress Report},
  author={Heinmiller, Jennifer KN},
  journal={Dictionaries: Journal of the Dictionary Society of North America},
  volume={44},
  number={1},
  pages={91--104},
  year={2023},
  publisher={Dictionary Society of North America}
}

@inproceedings{rombach2022high,
  title={High-resolution image synthesis with latent diffusion models},
  author={Rombach, Robin and Blattmann, Andreas and Lorenz, Dominik and Esser, Patrick and Ommer, Bj{\"o}rn},
  booktitle={Proceedings of the IEEE/CVF conference on computer vision and pattern recognition},
  pages={10684--10695},
  year={2022}
}

@article{podell2023sdxl,
  title={Sdxl: Improving latent diffusion models for high-resolution image synthesis},
  author={Podell, Dustin and English, Zion and Lacey, Kyle and Blattmann, Andreas and Dockhorn, Tim and M{\"u}ller, Jonas and Penna, Joe and Rombach, Robin},
  journal={arXiv preprint arXiv:2307.01952},
  year={2023}
}

@article{esser2024scaling,
  title={Scaling rectified flow transformers for high-resolution image synthesis, 2024},
  author={Esser, Patrick and Kulal, Sumith and Blattmann, Andreas and Entezari, Rahim and M{\"u}ller, Jonas and Saini, Harry and Levi, Yam and Lorenz, Dominik and Sauer, Axel and Boesel, Frederic and others},
  journal={URL https://arxiv. org/abs/2403.03206},
  volume={2},
  year={2024}
}

@article{wan2024survey,
  title={Survey of Bias In Text-to-Image Generation: Definition},
  author={Wan, Yixin and Subramonian, Arjun and Ovalle, Anaelia and Lin, Zongyu and Suvarna, Ashima and Chance, Christina and Bansal, Hritik and Pattichis, Rebecca and Chang, Kai-Wei},
  journal={Evaluation, and Mitigation},
  year={2024}
}

@book{crystal2003english,
  title     = {English as a Global Language},
  author    = {Crystal, David},
  year      = {2003},
  edition   = {2},
  publisher = {Cambridge University Press},
  address   = {Cambridge},
  isbn      = {9780521823470}
}

@book{aeni2021literature,
  title={A literature review of English Language Variation on Sociolinguistics},
  author={Aeni, Nur and Octaberlina, Like Raskova and Lubis, Nenni Dwi Aprianti and others},
  year={2021},
  publisher={OSF}
}

@inproceedings{Ziems2022VALUEUD,
  title={VALUE: Understanding Dialect Disparity in NLU},
  author={Caleb Ziems and Jiaao Chen and Camille Harris and Jessica Brooke Anderson and Diyi Yang},
  booktitle={Annual Meeting of the Association for Computational Linguistics},
  year={2022}
}

@article{Ziems2023MultiVALUEAF,
  title={Multi-VALUE: A Framework for Cross-Dialectal English NLP},
  author={Caleb Ziems and William B. Held and Jingfeng Yang and Diyi Yang},
  journal={ACL 2023},
  year={2023},
}

@article{lee2023holistic,
  title={Holistic evaluation of text-to-image models},
  author={Lee, Tony and Yasunaga, Michihiro and Meng, Chenlin and Mai, Yifan and Park, Joon Sung and Gupta, Agrim and Zhang, Yunzhi and Narayanan, Deepak and Teufel, Hannah and Bellagente, Marco and others},
  journal={Advances in Neural Information Processing Systems},
  volume={36},
  pages={69981--70011},
  year={2023}
}

@inproceedings{blodgett2018twitter,
  title={Twitter universal dependency parsing for African-American and mainstream American English},
  author={Blodgett, Su Lin and Wei, Johnny and O’Connor, Brendan},
  booktitle={Proceedings of the 56th Annual Meeting of the Association for Computational Linguistics (Volume 1: Long Papers)},
  pages={1415--1425},
  year={2018}
}

@article{gururangan2022whose,
  title={Whose language counts as high quality? Measuring language ideologies in text data selection},
  author={Gururangan, Suchin and Card, Dallas and Dreier, Sarah K and Gade, Emily K and Wang, Leroy Z and Wang, Zeyu and Zettlemoyer, Luke and Smith, Noah A},
  journal={arXiv preprint arXiv:2201.10474},
  year={2022}
}

@inproceedings{bender2021dangers,
  title={On the dangers of stochastic parrots: Can language models be too big?},
  author={Bender, Emily M and Gebru, Timnit and McMillan-Major, Angelina and Shmitchell, Shmargaret},
  booktitle={Proceedings of the 2021 ACM conference on fairness, accountability, and transparency},
  pages={610--623},
  year={2021}
}

@inproceedings{jorgensen2015challenges,
  title={Challenges of studying and processing dialects in social media},
  author={J{\o}rgensen, Anna and Hovy, Dirk and S{\o}gaard, Anders},
  booktitle={Proceedings of the workshop on noisy user-generated text},
  pages={9--18},
  year={2015}
}

@inproceedings{hovy2016social,
  title={The social impact of natural language processing},
  author={Hovy, Dirk and Spruit, Shannon L},
  booktitle={Proceedings of the 54th Annual Meeting of the Association for Computational Linguistics (Volume 2: Short Papers)},
  pages={591--598},
  year={2016}
}

@article{hurst2024gpt4o,
  title={GPT-4o System Card},
  author={Hurst, Aaron and Lerer, Adam and Goucher, Adam P and Perelman, Adam and Ramesh, Aditya and Clark, Aidan and Ostrow, AJ and Welihinda, Akila and Hayes, Alan and Radford, Alec and others},
  journal={arXiv preprint arXiv:2410.21276},
  year={2024}
}

@misc{flux2024,
    author={{Black Forest Labs}},
    title={FLUX},
    year={2024},
    howpublished={\url{https://github.com/black-forest-labs/flux}},
}

@misc{Dayma_DALL·E_Mini_2021,
      author = {Dayma, Boris and Patil, Suraj and Cuenca, Pedro and Saifullah, Khalid and Abraham, Tanishq and Lê Khac, Phúc and Melas, Luke and Ghosh, Ritobrata},
      doi = {10.5281/zenodo.5146400},
      month = {7},
      title = {DALL·E Mini},
      url = {https://github.com/borisdayma/dalle-mini},
      year = {2021}
}

@article{wan2025,
      title={Wan: Open and Advanced Large-Scale Video Generative Models}, 
      author={Ang Wang and Baole Ai and Bin Wen and Chaojie Mao and Chen-Wei Xie and Di Chen and Feiwu Yu and Haiming Zhao and Jianxiao Yang and Jianyuan Zeng and Jiayu Wang and Jingfeng Zhang and Jingren Zhou and Jinkai Wang and Jixuan Chen and Kai Zhu and Kang Zhao and Keyu Yan and Lianghua Huang and Mengyang Feng and Ningyi Zhang and Pandeng Li and Pingyu Wu and Ruihang Chu and Ruili Feng and Shiwei Zhang and Siyang Sun and Tao Fang and Tianxing Wang and Tianyi Gui and Tingyu Weng and Tong Shen and Wei Lin and Wei Wang and Wei Wang and Wenmeng Zhou and Wente Wang and Wenting Shen and Wenyuan Yu and Xianzhong Shi and Xiaoming Huang and Xin Xu and Yan Kou and Yangyu Lv and Yifei Li and Yijing Liu and Yiming Wang and Yingya Zhang and Yitong Huang and Yong Li and You Wu and Yu Liu and Yulin Pan and Yun Zheng and Yuntao Hong and Yupeng Shi and Yutong Feng and Zeyinzi Jiang and Zhen Han and Zhi-Fan Wu and Ziyu Liu},
      journal = {arXiv preprint arXiv:2503.20314},
      year={2025}
}

@article{yang2024cogvideox,
  title={Cogvideox: Text-to-video diffusion models with an expert transformer},
  author={Yang, Zhuoyi and Teng, Jiayan and Zheng, Wendi and Ding, Ming and Huang, Shiyu and Xu, Jiazheng and Yang, Yuanming and Hong, Wenyi and Zhang, Xiaohan and Feng, Guanyu and others},
  journal={arXiv preprint arXiv:2408.06072},
  year={2024}
}

@article{agarwal2025cosmos,
  title={Cosmos world foundation model platform for physical ai},
  author={Agarwal, Niket and Ali, Arslan and Bala, Maciej and Balaji, Yogesh and Barker, Erik and Cai, Tiffany and Chattopadhyay, Prithvijit and Chen, Yongxin and Cui, Yin and Ding, Yifan and others},
  journal={arXiv preprint arXiv:2501.03575},
  year={2025}
}

@article{betker2023improving,
  title={Improving image generation with better captions},
  author={Betker, James and Goh, Gabriel and Jing, Li and Brooks, Tim and Wang, Jianfeng and Li, Linjie and Ouyang, Long and Zhuang, Juntang and Lee, Joyce and Guo, Yufei and others},
  journal={Computer Science. https://cdn. openai. com/papers/dall-e-3. pdf},
  volume={2},
  number={3},
  pages={8},
  year={2023}
}

@misc{ramesh2022dalle2,
  author = {Ramesh, Aditya and Dhariwal, Prafulla and Nichol, Alex and Chu, Casey and Chen, Mark},
  title = {DALL·E 2: A New AI System that Can Create Realistic Images and Art from a Description in Natural Language},
  howpublished = {\url{https://openai.com/research/dall-e-2}},
  year = {2022}
}

@misc{openai2025_4o_image_generation,
  author = {OpenAI},
  title = {Introducing 4o Image Generation},
  howpublished = {\url{https://openai.com/index/introducing-4o-image-generation/}},
  year = {2025},
  note = {Accessed: 2025-05-19}
}

@article{grattafiori2024llama,
  title={The llama 3 herd of models},
  author={Grattafiori, Aaron and Dubey, Abhimanyu and Jauhri, Abhinav and Pandey, Abhinav and Kadian, Abhishek and Al-Dahle, Ahmad and Letman, Aiesha and Mathur, Akhil and Schelten, Alan and Vaughan, Alex and others},
  journal={arXiv preprint arXiv:2407.21783},
  year={2024}
}

@article{singh2025openai,
  title={Openai gpt-5 system card},
  author={Singh, Aaditya and Fry, Adam and Perelman, Adam and Tart, Adam and Ganesh, Adi and El-Kishky, Ahmed and McLaughlin, Aidan and Low, Aiden and Ostrow, AJ and Ananthram, Akhila and others},
  journal={arXiv preprint arXiv:2601.03267},
  year={2025}
}

@inproceedings{wallace2024diffusion,
  title={Diffusion model alignment using direct preference optimization},
  author={Wallace, Bram and Dang, Meihua and Rafailov, Rafael and Zhou, Linqi and Lou, Aaron and Purushwalkam, Senthil and Ermon, Stefano and Xiong, Caiming and Joty, Shafiq and Naik, Nikhil},
  booktitle={Proceedings of the IEEE/CVF Conference on Computer Vision and Pattern Recognition},
  pages={8228--8238},
  year={2024}
}

@article{dai2023emu,
  title={Emu: Enhancing image generation models using photogenic needles in a haystack},
  author={Dai, Xiaoliang and Hou, Ji and Ma, Chih-Yao and Tsai, Sam and Wang, Jialiang and Wang, Rui and Zhang, Peizhao and Vandenhende, Simon and Wang, Xiaofang and Dubey, Abhimanyu and others},
  journal={arXiv preprint arXiv:2309.15807},
  year={2023}
}

@article{Segalis2023API,
  title={A Picture is Worth a Thousand Words: Principled Recaptioning Improves Image Generation},
  author={Eyal Segalis and Dani Valevski and Danny Lumen and Yossi Matias and Yaniv Leviathan},
  journal={ArXiv},
  year={2023},
  volume={abs/2310.16656},
  url={https://api.semanticscholar.org/CorpusID:266003242}
}

@article{clark2023directly,
  title={Directly fine-tuning diffusion models on differentiable rewards},
  author={Clark, Kevin and Vicol, Paul and Swersky, Kevin and Fleet, David J},
  journal={arXiv preprint arXiv:2309.17400},
  year={2023}
}

@article{prabhudesai2023aligning,
  title={Aligning text-to-image diffusion models with reward backpropagation},
  author={Prabhudesai, Mihir and Goyal, Anirudh and Pathak, Deepak and Fragkiadaki, Katerina},
  journal={arXiv preprint arXiv:2310.03739},
  year={2023}
}

@article{black2023training,
  title={Training diffusion models with reinforcement learning},
  author={Black, Kevin and Janner, Michael and Du, Yilun and Kostrikov, Ilya and Levine, Sergey},
  journal={arXiv preprint arXiv:2305.13301},
  year={2023}
}

@article{fan2023dpok,
  title={Dpok: Reinforcement learning for fine-tuning text-to-image diffusion models},
  author={Fan, Ying and Watkins, Olivia and Du, Yuqing and Liu, Hao and Ryu, Moonkyung and Boutilier, Craig and Abbeel, Pieter and Ghavamzadeh, Mohammad and Lee, Kangwook and Lee, Kimin},
  journal={Advances in Neural Information Processing Systems},
  volume={36},
  pages={79858--79885},
  year={2023}
}

@article{dang2025personalized,
  title={Personalized Preference Fine-tuning of Diffusion Models},
  author={Dang, Meihua and Singh, Anikait and Zhou, Linqi and Ermon, Stefano and Song, Jiaming},
  journal={arXiv preprint arXiv:2501.06655},
  year={2025}
}

@article{hao2023optimizing,
  title={Optimizing prompts for text-to-image generation},
  author={Hao, Yaru and Chi, Zewen and Dong, Li and Wei, Furu},
  journal={Advances in Neural Information Processing Systems},
  volume={36},
  pages={66923--66939},
  year={2023}
}

@inproceedings{wang2024promptcharm,
  title={Promptcharm: Text-to-image generation through multi-modal prompting and refinement},
  author={Wang, Zhijie and Huang, Yuheng and Song, Da and Ma, Lei and Zhang, Tianyi},
  booktitle={Proceedings of the 2024 CHI Conference on Human Factors in Computing Systems},
  pages={1--21},
  year={2024}
}

@inproceedings{chen2024tailored,
  title={Tailored visions: Enhancing text-to-image generation with personalized prompt rewriting},
  author={Chen, Zijie and Zhang, Lichao and Weng, Fangsheng and Pan, Lili and Lan, Zhenzhong},
  booktitle={Proceedings of the IEEE/CVF Conference on Computer Vision and Pattern Recognition},
  pages={7727--7736},
  year={2024}
}

@inproceedings{lin2024evaluating,
  title={Evaluating text-to-visual generation with image-to-text generation},
  author={Lin, Zhiqiu and Pathak, Deepak and Li, Baiqi and Li, Jiayao and Xia, Xide and Neubig, Graham and Zhang, Pengchuan and Ramanan, Deva},
  booktitle={European Conference on Computer Vision},
  pages={366--384},
  year={2024},
  organization={Springer}
}

@inproceedings{Radford2021LearningTV,
  title={Learning Transferable Visual Models From Natural Language Supervision},
  author={Alec Radford and Jong Wook Kim and Chris Hallacy and Aditya Ramesh and Gabriel Goh and Sandhini Agarwal and Girish Sastry and Amanda Askell and Pamela Mishkin and Jack Clark and Gretchen Krueger and Ilya Sutskever},
  booktitle={International Conference on Machine Learning},
  year={2021},
}

@inproceedings{Hessel2021CLIPScoreAR,
  title={CLIPScore: A Reference-free Evaluation Metric for Image Captioning},
  author={Jack Hessel and Ari Holtzman and Maxwell Forbes and Ronan Joseph Le Bras and Yejin Choi},
  booktitle={Conference on Empirical Methods in Natural Language Processing},
  year={2021}
}

@inproceedings{zhou2025contrastive,
  title={Contrastive Visual Data Augmentation},
  author={Zhou, Yu and Li, Bingxuan and Tang, Mohan and Jin, Xiaomeng and Wu, Te-Lin and Huang, Kuan-Hao and Ji, Heng and Chang, Kai-Wei and Peng, Nanyun},
  booktitle={ICML 2025},
  year={2025}
}

@article{pearson1895,
  author  = {Pearson, Karl},
  title   = {Notes on regression and inheritance in the case of two parents},
  journal = {Proceedings of the Royal Society of London},
  volume  = {58},
  pages   = {240--242},
  year    = {1895},
  doi     = {10.1098/rspl.1895.0041}
}

@inproceedings{zhang2023adding,
  title={Adding conditional control to text-to-image diffusion models},
  author={Zhang, Lvmin and Rao, Anyi and Agrawala, Maneesh},
  booktitle={Proceedings of the IEEE/CVF international conference on computer vision},
  pages={3836--3847},
  year={2023}
}

@article{geng2024unmet,
  title={The unmet promise of synthetic training images: Using retrieved real images performs better},
  author={Geng, Scott and Hsieh, Cheng-Yu and Ramanujan, Vivek and Wallingford, Matthew and Li, Chun-Liang and Koh, Pang Wei W and Krishna, Ranjay},
  journal={Advances in Neural Information Processing Systems},
  volume={37},
  pages={7902--7929},
  year={2024}
}

@article{li2025real,
  title={Real: Realism evaluation of text-to-image generation models for effective data augmentation},
  author={Li, Ran and Jin, Xiaomeng and others},
  journal={arXiv preprint arXiv:2502.10663},
  year={2025}
}

@article{chung2023unimax,
  title={Unimax: Fairer and more effective language sampling for large-scale multilingual pretraining},
  author={Chung, Hyung Won and Constant, Noah and Garcia, Xavier and Roberts, Adam and Tay, Yi and Narang, Sharan and Firat, Orhan},
  journal={arXiv preprint arXiv:2304.09151},
  year={2023}
}

@article{ilharco2021openclip,
  title={Openclip},
  author={Ilharco, Gabriel and Wortsman, Mitchell and Carlini, Nicholas and Taori, Rohan and Dave, Achal and Shankar, Vaishaal and Namkoong, Hongseok and Miller, John and Hajishirzi, Hannaneh and Farhadi, Ali and others},
  journal={Zenodo},
  year={2021}
}

@article{schuhmann2022laion,
  title={Laion-5b: An open large-scale dataset for training next generation image-text models},
  author={Schuhmann, Christoph and Beaumont, Romain and Vencu, Richard and Gordon, Cade and Wightman, Ross and Cherti, Mehdi and Coombes, Theo and Katta, Aarush and Mullis, Clayton and Wortsman, Mitchell and others},
  journal={Advances in neural information processing systems},
  volume={35},
  pages={25278--25294},
  year={2022}
}
\bibliographystyle{colm2026_conference}
\clearpage
\appendix
\clearpage
\newpage
\appendix
\section*{Appendix}
\label{sec:appendix}

\section{Qualitative Comparison and Analysis of Mitigation Methods}
\label{subsec:qualitative_comparison}
\subsection{Decoder Fine-tuning}
\begin{figure*}[htbp]
\centering
    \includegraphics[width=1.0\linewidth]{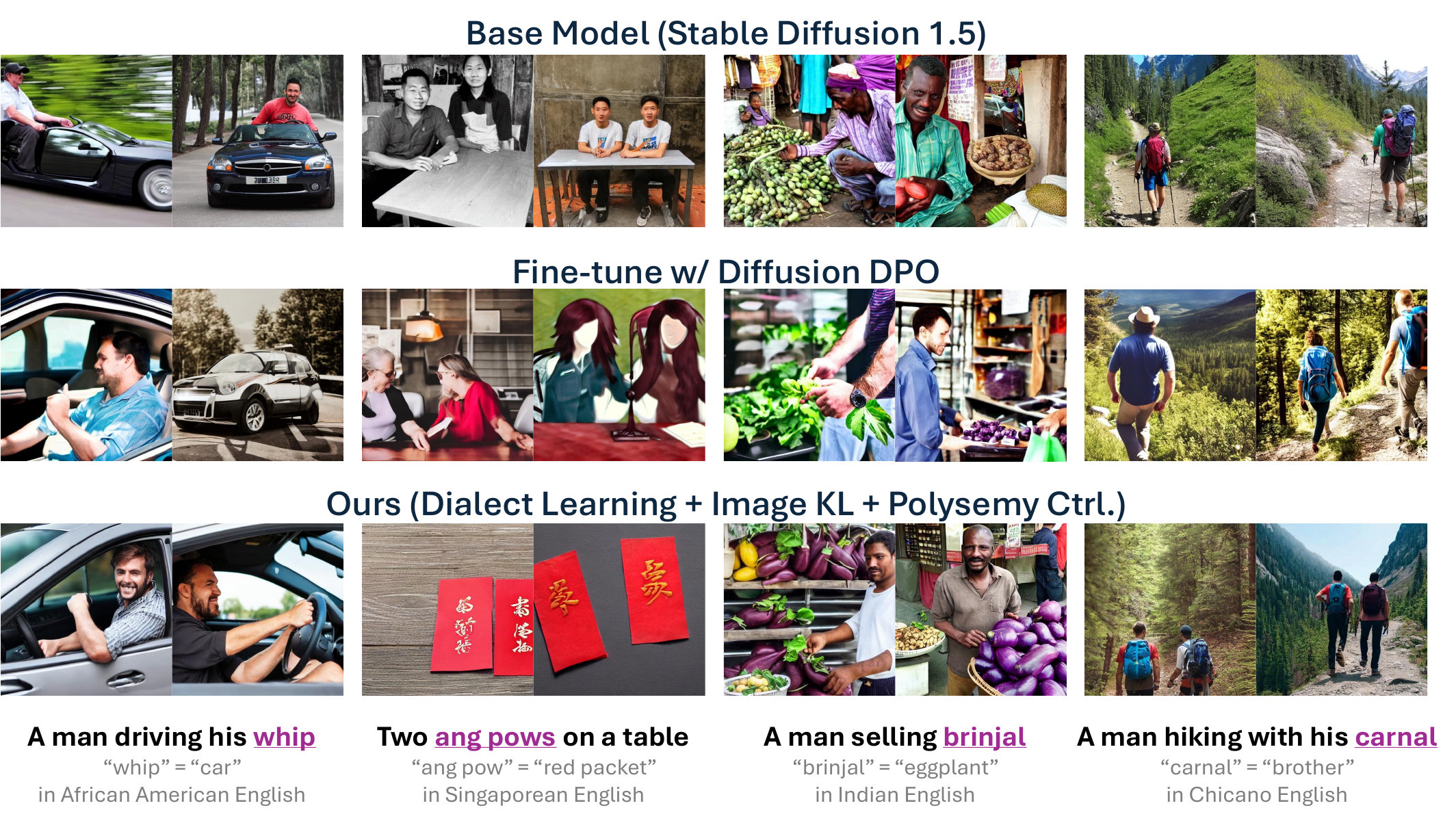}
    \caption{
    \textbf{Qualitative Comparison of Mitigation Strategies} using the Stable Diffusion 1.5 model~\citep{rombach2022high} on four different dialect prompts. Specifically, we compare the dialect prompt image generation results of the Stable Diffusion 1.5 Base Model, Stable Diffusion 1.5 fine-tuned with Diffusion DPO~\citep{wallace2024diffusion}, and Stable Diffusion 1.5 updated via our best performing method (Dialect Learning + Image KL Regularization + Polysemy Control).}
    \label{fig:qualitative}
\end{figure*}

In \Cref{fig:qualitative}, we provide additional qualitative examples to demonstrate the performances of the baseline mitigation strategy, Diffusion DPO~\citep{wallace2024diffusion}, compared with our method. Specifically, we update the Stable Diffusion 1.5 model encoder using Dialect Learning, Polysemy Control, and Image KL. After mitigation, we ask each model to generate images based on the four dialect prompts first mentioned in ~\Cref{fig:teaser}. The Stable Diffusion 1.5 Base model struggles to generate correct images for most of these prompts, including "Two ang pows on a table", "A man selling brinjal", and "A man hiking with his carnal". While the model is able to generate moderately reasonable images for the prompt "A man driving his whip", it commonly generates physically implausible details such as the man's torso protruding through the car.

Fine-tuning the decoder with Diffusion DPO is able to slightly improve generation alignment with the text prompt (\textit{e.g.}, occasionally generating two people for the prompt "A man hiking with his carnal"). However, it more often blends visual elements within the desired target images with other irrelevant objects (\textit{e.g.}, generating a man selling purple pastries in place of eggplants or a man wearing a purple shirt holding vegetables). Our method generates higher-quality and better aligned images compared to the base model and Diffusion DPO by accurately learning to generate the target concepts without negatively impacting image quality. A significant majority of images in our sampled generations are able to generate images that correctly depict the target prompts, in line with quantitative evaluation results.

\subsection{Prompt Revision}
\label{sec:appendix_prompt_revision}
\begin{table*}[t]

\centering
\resizebox{\textwidth}{!}{
\begin{tabular}{l|cc|cc|cc|cc|cc}
\toprule
\multirow{3}{*}{\textbf{Mitigation Methods}} & \multicolumn{10}{c}{\textbf{Performance by Dialect (VQAScore)} $\uparrow$} \\
\cmidrule(lr){2-11} 
& \multicolumn{2}{c|}{AAE} & \multicolumn{2}{c|}{BrE} & \multicolumn{2}{c|}{ChE} & \multicolumn{2}{c|}{InE} & \multicolumn{2}{c}{SgE} \\
& \textbf{Dialect} & \textbf{SAE} & \textbf{Dialect} & \textbf{SAE} & \textbf{Dialect} & \textbf{SAE} & \textbf{Dialect} & \textbf{SAE} & \textbf{Dialect} & \textbf{SAE} \\
\midrule
\textbf{Base Model (Stable Diffusion 1.5)} & \cellcolor{mygreen!16} 60.13 & \cellcolor{mygreen!36} 72.94 & \cellcolor{mygreen!31} 69.39 & \cellcolor{mygreen!42} 76.4 & \cellcolor{mygreen!4} 52.65 & \cellcolor{mygreen!45} 78.66 & 49.94 & \cellcolor{mygreen!49} 81.05 & \cellcolor{mygreen!11} 56.89 & \cellcolor{mygreen!48} 80.5 \\
\midrule
\textbf{Prompt Revision} &  &  &  &  &  &  &  &  &  &  \\
\hspace{1em}DALL-E 3 Prompt Rewrite & \cellcolor{mygreen!12} 57.73 & \cellcolor{mygreen!37} 73.16 & \cellcolor{mygreen!32} 70.4 & \cellcolor{mygreen!44} 77.86 & \cellcolor{mygreen!6} 53.98 & \cellcolor{mygreen!47} 79.99 & \cellcolor{mygreen!1} 50.42 & \cellcolor{mygreen!49} 81.33 & \cellcolor{mygreen!16} 59.87 & \cellcolor{mygreen!50} 81.66 \\
\hspace{1em}LLaMA 3 Prompt Translate & \cellcolor{mygreen!17} 60.87 & \cellcolor{mygreen!32} 70.36 & \cellcolor{mygreen!39} 74.39 & \cellcolor{mygreen!42} 76.49 & \cellcolor{mygreen!14} 59.05 & \cellcolor{mygreen!44} 78.15 & \cellcolor{mygreen!5} 53.04 & \cellcolor{mygreen!49} 81.09 & \cellcolor{mygreen!24} 64.98 & \cellcolor{mygreen!47} 79.84 \\
\hspace{1em}GPT 5 Prompt Translate & \cellcolor{mygreen!20} 62.35 & \cellcolor{mygreen!34} 71.45 & \cellcolor{mygreen!37} 73.18 & \cellcolor{mygreen!43} 77.03 & \cellcolor{mygreen!16} 59.91 & \cellcolor{mygreen!46} 79.02 & \cellcolor{mygreen!16} 60.22 & \cellcolor{mygreen!49} 80.75 & \cellcolor{mygreen!27} 67.33 & \cellcolor{mygreen!47} 79.67 \\
\textbf{Decoder Fine-tuning} &  &  &  &  &  &  &  &  &  &  \\
\hspace{1em}Diffusion Finetune & \cellcolor{mygreen!22} 63.85 & \cellcolor{mygreen!23} 64.34 & \cellcolor{mygreen!32} 70.14 & \cellcolor{mygreen!29} 68.35 & \cellcolor{mygreen!12} 57.3 & \cellcolor{mygreen!31} 69.55 & \cellcolor{mygreen!5} 52.84 & \cellcolor{mygreen!33} 70.72 & \cellcolor{mygreen!17} 60.56 & \cellcolor{mygreen!35} 72.42 \\
\hspace{1em}Diffusion DPO & \cellcolor{mygreen!26} 66.31 & \cellcolor{mygreen!21} 63.02 & \cellcolor{mygreen!30} 68.91 & \cellcolor{mygreen!30} 69.17 & \cellcolor{mygreen!18} 61.22 & \cellcolor{mygreen!28} 67.83 & \cellcolor{mygreen!10} 56.38 & \cellcolor{mygreen!33} 70.94 & \cellcolor{mygreen!23} 64.79 & \cellcolor{mygreen!35} 71.85 \\
\textbf{Ours} &  &  &  &  &  &  &  &  &  &  \\
\hspace{1em}Dialect Learning & \cellcolor{mygreen!40} 75.21 & \cellcolor{mygreen!38} 74.31 & \cellcolor{mygreen!45} 78.33 & \cellcolor{mygreen!45} 78.34 & \cellcolor{mygreen!46} 79.31 & \cellcolor{mygreen!48} 80.2 & \cellcolor{mygreen!44} 78.1 & \cellcolor{mygreen!47} 79.9 & \cellcolor{mygreen!46} 79.15 & \cellcolor{mygreen!45} 78.33 \\
\hspace{1.5em}+ Text Cosine Reg. & \cellcolor{mygreen!40} 75.44 & \cellcolor{mygreen!39} 74.86 & \cellcolor{mygreen!44} 77.84 & \cellcolor{mygreen!43} 77.52 & \cellcolor{mygreen!46} 79.31 & \cellcolor{mygreen!47} 79.74 & \cellcolor{mygreen!45} 78.22 & \cellcolor{mygreen!48} 80.13 & \cellcolor{mygreen!46} 78.86 & \cellcolor{mygreen!46} 79.21 \\
\hspace{1.5em}+ Image Cosine Reg. & \cellcolor{mygreen!39} 74.91 & \cellcolor{mygreen!39} 74.83 & \cellcolor{mygreen!45} 78.2 & \cellcolor{mygreen!45} 78.22 & \cellcolor{mygreen!47} 79.45 & \cellcolor{mygreen!48} 80.32 & \cellcolor{mygreen!44} 78 & \cellcolor{mygreen!47} 80 & \cellcolor{mygreen!46} 79.11 & \cellcolor{mygreen!45} 78.72 \\
\hspace{1.5em}+ Text KL Reg. & \cellcolor{mygreen!39} 74.4 & \cellcolor{mygreen!38} 73.97 & \cellcolor{mygreen!45} 78.27 & \cellcolor{mygreen!46} 79.4 & \cellcolor{mygreen!45} 78.36 & \cellcolor{mygreen!49} 80.72 & \cellcolor{mygreen!44} 78.17 & \cellcolor{mygreen!45} 78.24 & \cellcolor{mygreen!47} 79.71 & \cellcolor{mygreen!45} 78.66 \\
\hspace{1.5em}+ Image KL Reg. & \cellcolor{mygreen!38} 73.77 & \cellcolor{mygreen!38} 74.36 & \cellcolor{mygreen!43} 77.23 & \cellcolor{mygreen!44} 77.6 & \cellcolor{mygreen!46} 79.06 & \cellcolor{mygreen!48} 80.43 & \cellcolor{mygreen!46} 79.25 & \cellcolor{mygreen!49} 80.99 & \cellcolor{mygreen!49} 81.29 & \cellcolor{mygreen!47} 79.54 \\
\hspace{1.5em}+ Text KL Reg.+ Polysemy Ctrl. & \cellcolor{mygreen!35} 72.24 & \cellcolor{mygreen!35} 72.25 & \cellcolor{mygreen!41} 75.76 & \cellcolor{mygreen!47} 79.57 & \cellcolor{mygreen!46} 78.95 & \cellcolor{mygreen!46} 79.27 & \cellcolor{mygreen!48} 80.67 & \cellcolor{mygreen!47} 79.89 & \cellcolor{mygreen!49} 81.07 & \cellcolor{mygreen!47} 79.84 \\
\hspace{1.5em}+ Image KL Reg.+ Polysemy Ctrl. & \cellcolor{mygreen!36} 72.61 & \cellcolor{mygreen!38} 74.3 & \cellcolor{mygreen!42} 76.74 & \cellcolor{mygreen!42} 76.77 & \cellcolor{mygreen!43} 77.51 & \cellcolor{mygreen!46} 78.83 & \cellcolor{mygreen!48} 80.41 & \cellcolor{mygreen!49} 80.85 & \cellcolor{mygreen!49} 81.14 & \cellcolor{mygreen!44} 78.15 \\



 

\bottomrule
\end{tabular}
}

\caption{\textbf{Stable Diffusion 1.5 Mitigation Performance Breakdown} by dialect for different mitigation methods on the \Dataset{} dataset for all baseline methods and ablations of our method. All performance scores are measured using VQAScore~\citep{lin2024evaluating}, higher score is better. }
\label{tab:mitigation-methods-detail}
\end{table*}

As shown in ~\Cref{tab:mitigation-methods-detail}, ~\Cref{tab:mitigation-methods-sdxl}, and~\Cref{tab:mitigation-methods-wan21}, Prompt Revision methods yield only modest improvements in dialect performance (up to 6.8\% on Stable Diffusion 1.5), despite the fact that large language models such as GPT 5~\citep{singh2025openai} can often translate individual dialect sentences into Standard American English (SAE) with high accuracy in isolation. In this section, we provide a detailed qualitative error analysis on our experiments, showing why this standalone translation capability does not translate into proportional gains in end-to-end multimodal generation, and further discuss the practical deployment considerations for prompt translation and rewriting strategies.

\paragraph{Blind application at inference time.}
A fundamental limitation of prompt revision is that it must be applied uniformly to all incoming prompts during deployment. Since real-world systems cannot reliably detect whether a given prompt is written in a specific dialect or in SAE, the rewriting step must be applied indiscriminately. This introduces unnecessary semantic drift and added computational overhead even for prompts that are already in SAE. In state-of-the-art multimodal generative model systems and services, prompt revision techniques are used to remove harmful input and enrich aesthetic details, while less commonly used for translation due to the concern of injecting bias and misinterpretation at scale.

\paragraph{Mis-correction of valid dialect lexemes.}
A common failure mode occurs when the LLM misinterprets a dialect lexeme as a typographical error and ``corrects'' it to an unrelated SAE word. For example, given the dialect prompt \textit{``a photo of the matha''}, where \textit{matha} refers to a monastery in Indian English, the intended SAE meaning is \textit{``a photo of the monastery''}. However, the LLM instead produces \textit{``a photo of the math''}, treating \textit{matha} as a misspelling rather than a valid dialect term. This type of mis-correction results in a semantically incorrect prompt that leads the generative model further from the intended visual output.

\paragraph{Tone polishing without lexical translation.}
In other cases, the LLM refines the grammatical structure or tone of the prompt while leaving the critical dialect lexeme entirely untranslated. For instance, for the dialect prompt \textit{``a woman getting help from the dai''}, where \textit{dai} means \textit{midwife} in Indian English, the intended SAE meaning is \textit{``a woman receiving help from the midwife''}. The LLM instead produces \textit{``a woman receiving help from the dai''}, improving the phrasing of the surrounding context but failing to translate the key dialectal word. Because the dialect lexeme is preserved, the generative model still cannot correctly interpret the prompt, rendering the rewriting ineffective, while not harmful.

\paragraph{Polysemy and context-selection errors.}
When a dialect lexeme is polysemous, the LLM may select the wrong sense during translation, leading to systematically incorrect generations. For example, given the Singlish prompt \textit{``a guy feeling blur at the front of the class''}, where \textit{blur} is a common Singaporean English expression meaning \textit{confused}. The intended SAE meaning is \textit{``a guy feeling confused at the front of the class''}. However, the LLM selects the visual sense of the word and produces \textit{``a blurry guy at the front of the class''}, which causes the generative model to produce an image with blur artifacts applied to the person rather than depicting a confused student. This type of sense-selection error is particularly problematic because it produces a grammatically valid but semantically wrong prompt that the generative model faithfully renders.

\paragraph{Brittleness of concise prompts to paraphrasing.}
Concise prompts, which typically consist of six or fewer words, are especially vulnerable to the paraphrasing behavior of LLM-based rewriting systems. Even small modifications to such short prompts can significantly alter the visual content that generative models produce. We observe two specific sub-patterns of this failure mode.

\begin{itemize}[topsep=0.5em, itemsep=0em, leftmargin=15pt]
    \item \textbf{Over-augmentation}, where the LLM adds contextual details that were not present in the original prompt, steering the generative model toward unintended visual elements. For example, the Singaporean English prompt \textit{``two kids opening ang pow''} (where \textit{ang pow} means \textit{red packets}) should translate to \textit{``two kids opening red packets''}. Instead, the LLM produces \textit{``two happy children opening traditional red envelopes for Chinese New Year''}, injecting culture-specific context that biases the generative model toward producing scenes with lion dances, fireworks, and large crowds rather than the simple scene described in the original prompt.
    \item \textbf{Semantic drift}, where the LLM introduces descriptive modifiers that subtly shift the visual semantics. For instance, the Indian English prompt \textit{``a boy eating pani puri''} requires no translation, as \textit{pani puri} is already a well-known term for a specific Indian cuisine. However, the LLM rewrites it as \textit{``a boy enjoying a type of spicy Indian street food''}, where the added adjectives \textit{spicy} and \textit{street food} lead the generative model to depict crowded outdoor market stalls while missing the key element of \textit{pani puri} as intended by the user.
\end{itemize}

\section{Mitigation Results on Stable Diffusion XL and Wan 2.1 Video}
\label{subsec:additional_mitigation-results}

In main paper~\Cref{subsec:mitigation_experiments}, we discussed experiment results of common bias mitigation methods and compared them with our proposed approach. We also ran detailed ablation experiments to show the effects of each loss component. In this section, we will discuss mitigation results on two additional models: Stable Diffusion XL and Wan 2.1 Video.

\subsection{Stable Diffusion XL}
\begin{table*}[t]

\centering
\resizebox{\textwidth}{!}{
\begin{tabular}{l|ccc|cccccc} 
\toprule
\multirow{3}{*}{\textbf{Mitigation Methods}} & \multicolumn{3}{c|}{\textbf{Overall Performances $\uparrow$}} & \multicolumn{5}{c}{\textbf{Dialect Performance} $\uparrow$} \\
& SAE & SAE & Dialect & \multirow{2}{*}{AAE} & \multirow{2}{*}{BrE} & \multirow{2}{*}{ChE} & \multirow{2}{*}{InE} & \multirow{2}{*}{SgE} \\
& MSCOCO & Polysemy & Avg. & & & & & \\
\midrule


\textbf{Base Model (Stable Diffusion XL)} & \cellcolor{mygreen!50} 86.21 & \cellcolor{mygreen!50} 78.21 & 62.5 & \cellcolor{mygreen!7} 62.97 & \cellcolor{mygreen!24} 80.49 & 49.82 & 53.66 & 65.56 \\
\midrule
\textbf{Prompt Revision} &  &  &  &  &  &  &  &  \\
\hspace{1em}DALL-E 3 Prompt Rewrite & \cellcolor{mygreen!47} 85.36 & \cellcolor{mygreen!50} 78.01 & \cellcolor{mygreen!8} 66.49 & 59.93 & \cellcolor{mygreen!8} 77.92 & \cellcolor{mygreen!15} 60.61 & \cellcolor{mygreen!14} 63.62 & \cellcolor{mygreen!11} 70.39 \\
\hspace{1em}LLaMA 3 Prompt Translate & \cellcolor{mygreen!45} 84.72 & \cellcolor{mygreen!49} 77.6 & \cellcolor{mygreen!4} 64.19 & \cellcolor{mygreen!8} 63.74 & \cellcolor{mygreen!8} 77.93 & \cellcolor{mygreen!10} 57.4 & \cellcolor{mygreen!3} 56.09 & \cellcolor{mygreen!1} 65.8 \\
\hspace{1em}GPT 5 Prompt Translate & \cellcolor{mygreen!49} 85.96 & \cellcolor{mygreen!50} 78.03 & \cellcolor{mygreen!14} 69.47 & \cellcolor{mygreen!6} 62.69 & \cellcolor{mygreen!21} 81.13 & \cellcolor{mygreen!21} 64.89 & \cellcolor{mygreen!19} 67.2 & \cellcolor{mygreen!14} 71.46 \\
\textbf{Decoder Fine-tuning} &  &  &  &  &  &  &  &  \\
\hspace{1em}Decoder Finetune & 70.49 & \cellcolor{mygreen!4} 52.37 & \cellcolor{mygreen!6} 65.22 & \cellcolor{mygreen!12} 65.31 & 76.69 & \cellcolor{mygreen!14} 60.12 & \cellcolor{mygreen!6} 58.05 & \cellcolor{mygreen!1} 65.91 \\
\hspace{1em}Diffusion DPO & \cellcolor{mygreen!5} 72.03 & 50.29 & \cellcolor{mygreen!9} 66.89 & \cellcolor{mygreen!13} 65.97 & \cellcolor{mygreen!9} 78.12 & \cellcolor{mygreen!18} 62.88 & \cellcolor{mygreen!9} 60.1 & \cellcolor{mygreen!4} 67.4 \\
\textbf{Ours} &  &  &  &  &  &  &  &  \\
\hspace{1em} Dialect Learning + Text KL Reg.+ Polysemy Reg. & \cellcolor{mygreen!48} 85.45 & \cellcolor{mygreen!50} 78.08 & \cellcolor{mygreen!50} 85.99 & \cellcolor{mygreen!50} 82.43 & \cellcolor{mygreen!50} 84.71 & \cellcolor{mygreen!50} 85.97 & \cellcolor{mygreen!50} 89.7 & \cellcolor{mygreen!50} 87.14 \\

\bottomrule
\end{tabular}
}

\caption{
\textbf{Mitigation results on SDXL~\citep{podell2023sdxl}} for all methods, including \textbf{Overall Performances} on SAE MSCOCO, SAE Polysemy, average Dialect performance, and \textbf{Dialect Performance} for each dialect, all measured using VQAScore~\citep{lin2024evaluating}. Cell colors reflect column-normalized performance values, with darker green indicating higher VQAScore performance.
}
\label{tab:mitigation-methods-sdxl}
\end{table*}
Stable Diffusion XL consists of two encoders: a Base encoder and a Refiner encoder. We fine-tuned both components as part of our method. However, since the corresponding CLIP-style image encoder for the Refiner is not publicly accessible, only Text KL Regularization can be applied in this case. Given the Refiner's larger size and additional encoding modules, we evaluate our final method against other baselines within this more complex configuration.

We report the mitigation results on Stable Diffusion XL~\citep{podell2023sdxl} in \Cref{tab:mitigation-methods-sdxl}, under the experimental setup described above. Similar to the findings on Stable Diffusion 1.5, Prompt Revision methods preserve general SAE performance but yield only marginal improvements in dialect VQAScore, with gains of up to 7.8\%. Additionally, decoder fine-tuning methods also result in small gains of up to 5.3\% in dialect performance, but at the cost of noticeable degradation in both SAE MSCOCO and SAE polysemy performance. In contrast, our method substantially improves dialect robustness across all five dialects, achieving an average performance of 85.99\%, which surpasses the base model’s SAE score of 84.43\%, while inducing less than a 1 point drop in both SAE MSCOCO and SAE polysemy performance.

\subsection{Wan 2.1 Video}
\begin{table*}[t]

\centering
\resizebox{\textwidth}{!}{
\begin{tabular}{l|ccc|cccccc} 
\toprule
\multirow{3}{*}{\textbf{Mitigation Methods}} & \multicolumn{3}{c|}{\textbf{Overall Performances $\uparrow$}} & \multicolumn{5}{c}{\textbf{Dialect Performance} $\uparrow$} \\
& SAE & SAE & Dialect & \multirow{2}{*}{AAE} & \multirow{2}{*}{BrE} & \multirow{2}{*}{ChE} & \multirow{2}{*}{InE} & \multirow{2}{*}{SgE} \\
& MSCOCO & Polysemy & Avg. & & & & & \\
\midrule

\textbf{Base Model (Wan 2.1)} & \cellcolor{mygreen!50} 77.73 & \cellcolor{mygreen!50} 71.29 & 51.58 & \cellcolor{mygreen!4} 55.79 & \cellcolor{mygreen!3} 62.14 & 42.39 & 48.86 & \cellcolor{mygreen!5} 48.73 \\
\midrule
\textbf{Prompt Revision} & & & & & & & & \\
\hspace{1em}DALL-E 3 Prompt Rewrite         & \cellcolor{mygreen!47} 76.3 & \cellcolor{mygreen!50} 71.03 & \cellcolor{mygreen!3} 53.19 & 53.6 & 61.57 & \cellcolor{mygreen!9} 48.12 & \cellcolor{mygreen!9} 53.49 & \cellcolor{mygreen!6} 49.17 \\
\hspace{1em}LLaMA 3 Prompt Translate        & \cellcolor{mygreen!45} 75.48 & \cellcolor{mygreen!48} 69.98 & \cellcolor{mygreen!3} 53.06 & \cellcolor{mygreen!5} 55.97 & \cellcolor{mygreen!6} 62.81 & \cellcolor{mygreen!8} 47.02 & \cellcolor{mygreen!5} 51.43 & \cellcolor{mygreen!4} 48.06 \\
\hspace{1em}GPT 5 Prompt Translate         & \cellcolor{mygreen!49} 77.12 & \cellcolor{mygreen!49} 70.85 & \cellcolor{mygreen!14} 58 & \cellcolor{mygreen!8} 57.92 & \cellcolor{mygreen!30} 68.17 & \cellcolor{mygreen!19} 54.35 & \cellcolor{mygreen!16} 57.12 & \cellcolor{mygreen!11} 52.46 \\
\textbf{Decoder Fine-tuning}& & & & & & & & \\
\hspace{1em}Decoder Finetune              & 54.18 & 43.66 & \cellcolor{mygreen!7} 54.81 & \cellcolor{mygreen!9} 58.13 & \cellcolor{mygreen!7} 63.19 & \cellcolor{mygreen!13} 50.12 & \cellcolor{mygreen!15} 56.71 & 45.89 \\
\textbf{Ours} & & & & & & & & \\
\hspace{1em} Dialect Learning + Text KL Reg.+ Polysemy Reg. & \cellcolor{mygreen!48} 76.89 & \cellcolor{mygreen!50} 71.12 & \cellcolor{mygreen!50} 75.02 & \cellcolor{mygreen!50} 79.13 & \cellcolor{mygreen!50} 72.69 & \cellcolor{mygreen!50} 73.1 & \cellcolor{mygreen!50} 75.19 & \cellcolor{mygreen!50} 75.01 \\

\bottomrule
\end{tabular}
}

\caption{
\textbf{Mitigation results on Wan2.1~\citep{wan2025}} for all methods, including \textbf{Overall Performances} on SAE MSCOCO, SAE Polysemy, average Dialect performance, and \textbf{Dialect Performance} for each dialect, all measured using VQAScore~\citep{lin2024evaluating}. Cell colors reflect column-normalized performance values, with darker green indicating higher VQAScore performance.
}

\label{tab:mitigation-methods-wan21}
\end{table*}
Wan 2.1 Video consists of a single multilingual-pretrained text encoder (UniMax T5 XXL~\citep{chung2023unimax}), which is significantly larger in scale compared to CLIP-style encoders~\citep{radford2021learning, ilharco2021openclip} used by Stable Diffusion models. Furthermore, as detailed in \Cref{subsec:models_details}, video generation inference costs for Wan 2.1 Video are extremely high. Given our relatively limited compute budget for video generation evaluation, we also compare our final method against other baselines.

We report the mitigation results on Wan 2.1 Video~\citep{wan2025} in \Cref{tab:mitigation-methods-wan21}, under the experimental setup described above. Here Prompt Revision methods yield limited improvements in overall dialect performance, with gains of up to 6.4\%. Decoder fine-tuning also result in small gains of up to 3.2\% in dialect performance, and at the cost of noticeable degradation in both SAE MSCOCO and SAE polysemy performance. The Diffusion DPO algorithm is not directly applicable for video generative models and is therefore not tested in this experiment. Finally, our proposed mitigation method substantially improves dialect robustness across all five dialects, achieving an average performance gain of 23.44\%, while inducing less than a 1\% drop in both SAE MSCOCO and SAE polysemy performance.

\section{Complete \Dataset{} Model Performance Results}
\begin{table*}[t!]

\centering
\resizebox{\textwidth}{!}{
\begin{tabular}{ccl|cc|cc|cc|cc|cc}
\toprule
\multicolumn{2}{c}{} & \multirow{3}{*}{\textbf{Model}} & \multicolumn{10}{c}{\textbf{Performance by Dialect (VQAScore) $\uparrow$}} \\
\cmidrule(lr){4-13} 
\multicolumn{2}{c}{} & & \multicolumn{2}{c|}{AAE} & \multicolumn{2}{c|}{BrE} & \multicolumn{2}{c|}{ChE} & \multicolumn{2}{c|}{InE} & \multicolumn{2}{c}{SgE} \\
\multicolumn{2}{c}{} & & \textbf{Dialect} & \textbf{SAE} & \textbf{Dialect} & \textbf{SAE} & \textbf{Dialect} & \textbf{SAE} & \textbf{Dialect} & \textbf{SAE} & \textbf{Dialect} & \textbf{SAE} \\
\midrule
\multirow{16}{*}{\rotatebox[origin=c]{90}{\textbf{Concise Prompts}}}
& \multirow{10}{*}{\rotatebox[origin=c]{90}{\textit{T2I Models}}}

& Stable Diffusion 1.4 & \cellcolor{mygreen!26} 60.66 & \cellcolor{mygreen!38} 76.47 & \cellcolor{mygreen!34} 71.46 & \cellcolor{mygreen!40} 79.08 & \cellcolor{mygreen!19} 51.31 & \cellcolor{mygreen!39} 78.86 & \cellcolor{mygreen!16} 47.5 & \cellcolor{mygreen!41} 80.88 & \cellcolor{mygreen!24} 57.64 & \cellcolor{mygreen!39} 78.92 \\
 &  & Stable Diffusion 1.5 & \cellcolor{mygreen!27} 62.31 & \cellcolor{mygreen!38} 77.41 & \cellcolor{mygreen!35} 72.59 & \cellcolor{mygreen!40} 79.47 & \cellcolor{mygreen!18} 50.4 & \cellcolor{mygreen!40} 79.37 & \cellcolor{mygreen!16} 47.03 & \cellcolor{mygreen!41} 81.29 & \cellcolor{mygreen!23} 56.36 & \cellcolor{mygreen!39} 78.8 \\
 &  & Stable Diffusion 2.1 & \cellcolor{mygreen!26} 60.97 & \cellcolor{mygreen!41} 80.59 & \cellcolor{mygreen!38} 76.37 & \cellcolor{mygreen!43} 84.21 & \cellcolor{mygreen!15} 45.88 & \cellcolor{mygreen!43} 83.15 & \cellcolor{mygreen!19} 50.63 & \cellcolor{mygreen!45} 85.99 & \cellcolor{mygreen!24} 58.53 & \cellcolor{mygreen!42} 82.31 \\
 &  & Stable Diffusion XL & \cellcolor{mygreen!28} 62.97 & \cellcolor{mygreen!42} 82.17 & \cellcolor{mygreen!41} 80.49 & \cellcolor{mygreen!46} 87.44 & \cellcolor{mygreen!18} 49.82 & \cellcolor{mygreen!44} 84.75 & \cellcolor{mygreen!21} 53.66 & \cellcolor{mygreen!46} 87.6 & \cellcolor{mygreen!30} 65.56 & \cellcolor{mygreen!43} 84.23 \\
 &  & Stable Diffusion 3 & \cellcolor{mygreen!26} 60.9 & \cellcolor{mygreen!44} 84.46 & \cellcolor{mygreen!40} 79.22 & \cellcolor{mygreen!45} 86.71 & \cellcolor{mygreen!17} 48.32 & \cellcolor{mygreen!43} 84.29 & \cellcolor{mygreen!19} 51.91 & \cellcolor{mygreen!46} 87.52 & \cellcolor{mygreen!27} 61.64 & \cellcolor{mygreen!42} 82.32 \\
 &  & Stable Diffusion 3.5 Large & \cellcolor{mygreen!26} 60.16 & \cellcolor{mygreen!43} 83.91 & \cellcolor{mygreen!41} 80.53 & \cellcolor{mygreen!47} 89.22 & \cellcolor{mygreen!17} 48.93 & \cellcolor{mygreen!44} 85.33 & \cellcolor{mygreen!19} 51.53 & \cellcolor{mygreen!47} 88.69 & \cellcolor{mygreen!28} 63.21 & \cellcolor{mygreen!43} 83.79 \\
 &  & Stable Diffusion 3.5 Large Turbo & \cellcolor{mygreen!23} 57.27 & \cellcolor{mygreen!42} 82.2 & \cellcolor{mygreen!40} 79.4 & \cellcolor{mygreen!46} 87.51 & \cellcolor{mygreen!16} 47.16 & \cellcolor{mygreen!43} 83.62 & \cellcolor{mygreen!18} 50.07 & \cellcolor{mygreen!45} 87.06 & \cellcolor{mygreen!27} 61.72 & \cellcolor{mygreen!43} 83.09 \\
 &  & Flux.1 [dev] & \cellcolor{mygreen!22} 55.63 & \cellcolor{mygreen!40} 80.17 & \cellcolor{mygreen!35} 72.7 & \cellcolor{mygreen!41} 81.53 & \cellcolor{mygreen!15} 45.85 & \cellcolor{mygreen!42} 82.82 & \cellcolor{mygreen!16} 46.73 & \cellcolor{mygreen!41} 81.39 & \cellcolor{mygreen!19} 51.63 & \cellcolor{mygreen!38} 76.62 \\
 &  & DALL-E Mini & \cellcolor{mygreen!19} 50.86 & \cellcolor{mygreen!38} 76.96 & \cellcolor{mygreen!35} 73.55 & \cellcolor{mygreen!40} 80.1 & \cellcolor{mygreen!12} 41.51 & \cellcolor{mygreen!39} 78.48 & \cellcolor{mygreen!14} 44.07 & \cellcolor{mygreen!38} 77.11 & \cellcolor{mygreen!21} 54.11 & \cellcolor{mygreen!35} 72.64 \\
 &  & DALL-E 2 & \cellcolor{mygreen!20} 52.07 & \cellcolor{mygreen!41} 81.19 & \cellcolor{mygreen!40} 79.19 & \cellcolor{mygreen!45} 86.03 & \cellcolor{mygreen!13} 42.54 & \cellcolor{mygreen!42} 83.05 & \cellcolor{mygreen!13} 43.11 & \cellcolor{mygreen!41} 81.66 & \cellcolor{mygreen!27} 61.65 & \cellcolor{mygreen!41} 81.27 \\
 &  & DALL-E 3 & \cellcolor{mygreen!31} 67.09 & \cellcolor{mygreen!42} 82.8 & \cellcolor{mygreen!44} 85.68 & \cellcolor{mygreen!47} 88.86 & \cellcolor{mygreen!18} 50.43 & \cellcolor{mygreen!45} 86.87 & \cellcolor{mygreen!25} 58.8 & \cellcolor{mygreen!45} 86.34 & \cellcolor{mygreen!29} 64.3 & \cellcolor{mygreen!45} 86.38 \\
 &  & DALL-E 3 w/ Rewrite & \cellcolor{mygreen!28} 63.74 & \cellcolor{mygreen!42} 81.83 & \cellcolor{mygreen!43} 84.24 & \cellcolor{mygreen!48} 90.08 & \cellcolor{mygreen!27} 61.41 & \cellcolor{mygreen!43} 83.96 & \cellcolor{mygreen!32} 68.7 & \cellcolor{mygreen!47} 89.28 & \cellcolor{mygreen!36} 74.77 & \cellcolor{mygreen!44} 85.69 \\
 &  & gpt-image-1 & \cellcolor{mygreen!29} 65.47 & \cellcolor{mygreen!47} 88.62 & \cellcolor{mygreen!46} 88.39 & \cellcolor{mygreen!50} 93.24 & \cellcolor{mygreen!29} 65.31 & \cellcolor{mygreen!46} 88.37 & \cellcolor{mygreen!31} 67.77 & \cellcolor{mygreen!49} 92.22 & \cellcolor{mygreen!39} 77.67 & \cellcolor{mygreen!46} 88.25 \\

\cmidrule(lr){2-13} 
& \multirow{5}{*}{\rotatebox[origin=c]{90}{\textit{T2V Models}}}

& Cosmos-1 & \cellcolor{mygreen!25} 59.61 & \cellcolor{mygreen!38} 76.57 & \cellcolor{mygreen!32} 68.87 & \cellcolor{mygreen!37} 76.26 & \cellcolor{mygreen!20} 53.27 & \cellcolor{mygreen!34} 72.08 & \cellcolor{mygreen!23} 56.84 & \cellcolor{mygreen!39} 78.34 & \cellcolor{mygreen!21} 54.04 & \cellcolor{mygreen!29} 65.18 \\
 &  & Open-Sora & \cellcolor{mygreen!29} 65.46 & \cellcolor{mygreen!44} 84.56 & \cellcolor{mygreen!37} 75.56 & \cellcolor{mygreen!43} 83.21 & \cellcolor{mygreen!17} 48.49 & \cellcolor{mygreen!44} 85.21 & \cellcolor{mygreen!25} 59.79 & \cellcolor{mygreen!46} 87.59 & \cellcolor{mygreen!25} 59.19 & \cellcolor{mygreen!41} 80.56 \\
 &  & VideoCrafter-2 & \cellcolor{mygreen!26} 61.3 & \cellcolor{mygreen!42} 82.13 & \cellcolor{mygreen!37} 76.19 & \cellcolor{mygreen!43} 84.12 & \cellcolor{mygreen!13} 42.9 & \cellcolor{mygreen!45} 86.43 & \cellcolor{mygreen!21} 53.3 & \cellcolor{mygreen!47} 88.76 & \cellcolor{mygreen!27} 61.73 & \cellcolor{mygreen!43} 83.51 \\
 &  & CogVideoX & \cellcolor{mygreen!8} 36.72 & \cellcolor{mygreen!25} 59.54 & \cellcolor{mygreen!13} 42.55 & \cellcolor{mygreen!22} 55.8 & \cellcolor{mygreen!2} 27.71 & \cellcolor{mygreen!27} 61.82 & \cellcolor{mygreen!2} 28.76 & \cellcolor{mygreen!28} 63.23 & 25.98 & \cellcolor{mygreen!14} 44 \\
 &  & Wan 2.1 & \cellcolor{mygreen!3} 29.57 & \cellcolor{mygreen!27} 62.49 & \cellcolor{mygreen!16} 47.02 & \cellcolor{mygreen!32} 68.41 & \cellcolor{mygreen!4} 30.37 & \cellcolor{mygreen!21} 54.07 & \cellcolor{mygreen!4} 30.68 & \cellcolor{mygreen!30} 65.81 & \cellcolor{mygreen!3} 30.23 & \cellcolor{mygreen!31} 67.89 \\

\midrule
\multirow{16}{*}{\rotatebox[origin=c]{90}{\textbf{Detailed Prompts}}}
& \multirow{10}{*}{\rotatebox[origin=c]{90}{\textit{T2I Models}}}

& Stable Diffusion 1.4 & \cellcolor{mygreen!33} 70.07 & \cellcolor{mygreen!40} 79.31 & \cellcolor{mygreen!36} 74.19 & \cellcolor{mygreen!38} 77.58 & \cellcolor{mygreen!29} 65.24 & \cellcolor{mygreen!39} 78.94 & \cellcolor{mygreen!23} 56.99 & \cellcolor{mygreen!41} 80.53 & \cellcolor{mygreen!28} 63.87 & \cellcolor{mygreen!38} 76.98 \\
 &  & Stable Diffusion 1.5 & \cellcolor{mygreen!34} 71.03 & \cellcolor{mygreen!40} 79.97 & \cellcolor{mygreen!35} 73.5 & \cellcolor{mygreen!39} 77.69 & \cellcolor{mygreen!29} 65.21 & \cellcolor{mygreen!39} 78.89 & \cellcolor{mygreen!23} 56.84 & \cellcolor{mygreen!40} 79.72 & \cellcolor{mygreen!28} 63.02 & \cellcolor{mygreen!38} 77.06 \\
 &  & Stable Diffusion XL & \cellcolor{mygreen!35} 72.82 & \cellcolor{mygreen!44} 84.76 & \cellcolor{mygreen!41} 80.84 & \cellcolor{mygreen!44} 85.6 & \cellcolor{mygreen!32} 68.19 & \cellcolor{mygreen!45} 85.85 & \cellcolor{mygreen!26} 61.1 & \cellcolor{mygreen!46} 87.44 & \cellcolor{mygreen!34} 70.93 & \cellcolor{mygreen!43} 83.55 \\
 &  & Stable Diffusion 2.1 & \cellcolor{mygreen!32} 69.41 & \cellcolor{mygreen!41} 81.72 & \cellcolor{mygreen!38} 77.51 & \cellcolor{mygreen!42} 82.03 & \cellcolor{mygreen!28} 63.64 & \cellcolor{mygreen!42} 82.68 & \cellcolor{mygreen!25} 59.39 & \cellcolor{mygreen!43} 84.07 & \cellcolor{mygreen!29} 64.71 & \cellcolor{mygreen!40} 79.95 \\
 &  & Stable Diffusion 3 & \cellcolor{mygreen!36} 74.27 & \cellcolor{mygreen!45} 87.11 & \cellcolor{mygreen!42} 82.58 & \cellcolor{mygreen!46} 88.48 & \cellcolor{mygreen!30} 66.21 & \cellcolor{mygreen!45} 86.95 & \cellcolor{mygreen!27} 62.59 & \cellcolor{mygreen!46} 88.08 & \cellcolor{mygreen!31} 67.32 & \cellcolor{mygreen!43} 83.13 \\
 &  & Stable Diffusion 3.5 Large & \cellcolor{mygreen!35} 73.21 & \cellcolor{mygreen!45} 86.84 & \cellcolor{mygreen!43} 83.24 & \cellcolor{mygreen!47} 89.5 & \cellcolor{mygreen!31} 67.05 & \cellcolor{mygreen!46} 87.6 & \cellcolor{mygreen!26} 60.55 & \cellcolor{mygreen!47} 88.82 & \cellcolor{mygreen!31} 67.65 & \cellcolor{mygreen!43} 84.27 \\
 &  & Stable Diffusion 3.5 Large Turbo & \cellcolor{mygreen!35} 73.24 & \cellcolor{mygreen!45} 86.23 & \cellcolor{mygreen!41} 81.07 & \cellcolor{mygreen!46} 88.24 & \cellcolor{mygreen!29} 64.83 & \cellcolor{mygreen!45} 86.37 & \cellcolor{mygreen!24} 58.46 & \cellcolor{mygreen!46} 87.81 & \cellcolor{mygreen!29} 65.05 & \cellcolor{mygreen!42} 82.98 \\
 &  & Flux.1 [dev] & \cellcolor{mygreen!35} 72.86 & \cellcolor{mygreen!44} 85.56 & \cellcolor{mygreen!38} 77.43 & \cellcolor{mygreen!44} 85.19 & \cellcolor{mygreen!27} 61.47 & \cellcolor{mygreen!42} 82.72 & \cellcolor{mygreen!24} 58.52 & \cellcolor{mygreen!44} 85.31 & \cellcolor{mygreen!25} 59.56 & \cellcolor{mygreen!40} 79.66 \\
 &  & DALL-E Mini & \cellcolor{mygreen!21} 53.69 & \cellcolor{mygreen!36} 74.12 & \cellcolor{mygreen!32} 69.5 & \cellcolor{mygreen!35} 73.38 & \cellcolor{mygreen!20} 52.39 & \cellcolor{mygreen!34} 72.11 & \cellcolor{mygreen!18} 50.47 & \cellcolor{mygreen!36} 73.65 & \cellcolor{mygreen!24} 58.22 & \cellcolor{mygreen!32} 68.92 \\
 &  & DALL-E 2 & \cellcolor{mygreen!29} 64.72 & \cellcolor{mygreen!40} 79.34 & \cellcolor{mygreen!40} 80.2 & \cellcolor{mygreen!45} 85.79 & \cellcolor{mygreen!27} 62.33 & \cellcolor{mygreen!43} 83.66 & \cellcolor{mygreen!22} 55.51 & \cellcolor{mygreen!42} 82.6 & \cellcolor{mygreen!30} 66.07 & \cellcolor{mygreen!40} 80.34 \\
 &  & DALL-E 3 & \cellcolor{mygreen!39} 77.75 & \cellcolor{mygreen!44} 85.3 & \cellcolor{mygreen!43} 83.82 & \cellcolor{mygreen!46} 87.99 & \cellcolor{mygreen!31} 68.16 & \cellcolor{mygreen!45} 86.26 & \cellcolor{mygreen!34} 71.19 & \cellcolor{mygreen!46} 87.79 & \cellcolor{mygreen!35} 73.51 & \cellcolor{mygreen!43} 84.35 \\
 &  & DALL-E 3 w/ Rewrite & \cellcolor{mygreen!38} 76.73 & \cellcolor{mygreen!45} 87.12 & \cellcolor{mygreen!44} 85.56 & \cellcolor{mygreen!48} 90.33 & \cellcolor{mygreen!38} 76.36 & \cellcolor{mygreen!44} 85.43 & \cellcolor{mygreen!37} 75.63 & \cellcolor{mygreen!49} 91.22 & \cellcolor{mygreen!39} 78.8 & \cellcolor{mygreen!45} 86.54 \\
 &  & gpt-image-1 & \cellcolor{mygreen!39} 78.26 & \cellcolor{mygreen!48} 90.7 & \cellcolor{mygreen!45} 86.88 & \cellcolor{mygreen!48} 90.94 & \cellcolor{mygreen!40} 79.47 & \cellcolor{mygreen!47} 88.85 & \cellcolor{mygreen!39} 78.04 & \cellcolor{mygreen!50} 92.86 & \cellcolor{mygreen!40} 79.39 & \cellcolor{mygreen!46} 88.38 \\

\cmidrule(lr){2-13} 
& \multirow{5}{*}{\rotatebox[origin=c]{90}{\textit{T2V Models}}}

& Cosmos-1 & \cellcolor{mygreen!29} 64.61 & \cellcolor{mygreen!35} 72.64 & \cellcolor{mygreen!31} 67.1 & \cellcolor{mygreen!36} 73.94 & \cellcolor{mygreen!24} 57.62 & \cellcolor{mygreen!31} 67.03 & \cellcolor{mygreen!23} 56.58 & \cellcolor{mygreen!35} 73 & \cellcolor{mygreen!18} 50.39 & \cellcolor{mygreen!25} 58.99 \\
 &  & Open-Sora & \cellcolor{mygreen!36} 74.81 & \cellcolor{mygreen!45} 86.48 & \cellcolor{mygreen!38} 76.69 & \cellcolor{mygreen!41} 80.84 & \cellcolor{mygreen!31} 67.65 & \cellcolor{mygreen!43} 83.93 & \cellcolor{mygreen!33} 69.59 & \cellcolor{mygreen!45} 86.77 & \cellcolor{mygreen!34} 71.15 & \cellcolor{mygreen!41} 81.49 \\
 &  & VideoCrafter-2 & \cellcolor{mygreen!33} 70.88 & \cellcolor{mygreen!44} 85.37 & \cellcolor{mygreen!40} 79.53 & \cellcolor{mygreen!42} 83 & \cellcolor{mygreen!30} 66.14 & \cellcolor{mygreen!46} 87.21 & \cellcolor{mygreen!27} 62.58 & \cellcolor{mygreen!45} 86.47 & \cellcolor{mygreen!31} 68.14 & \cellcolor{mygreen!43} 83.72 \\
 &  & CogVideoX & \cellcolor{mygreen!11} 39.83 & \cellcolor{mygreen!19} 50.63 & \cellcolor{mygreen!15} 46.4 & \cellcolor{mygreen!21} 54.35 & \cellcolor{mygreen!10} 38.89 & \cellcolor{mygreen!24} 57.82 & \cellcolor{mygreen!8} 35.8 & \cellcolor{mygreen!27} 62.68 & 25.51 & \cellcolor{mygreen!11} 40.11 \\
 &  & Wan 2.1 & \cellcolor{mygreen!22} 55.79 & \cellcolor{mygreen!40} 79.96 & \cellcolor{mygreen!27} 62.14 & \cellcolor{mygreen!35} 73.08 & \cellcolor{mygreen!12} 42.39 & \cellcolor{mygreen!36} 73.82 & \cellcolor{mygreen!17} 48.86 & \cellcolor{mygreen!38} 76.6 & \cellcolor{mygreen!17} 48.73 & \cellcolor{mygreen!37} 75.8 \\

\bottomrule
\end{tabular}
}

\caption{\textbf{Complete \Dataset{} Benchmark Performance Breakdown} by dialect for all text-to-image and text-to-video generative models. All performance scores are measured using VQAScore~\citep{lin2024evaluating}, higher score is better. Results complement \Cref{tab:main-benchmark-results} in the main paper.}
\label{tab:benchmark-results-vqa} 
\end{table*}
\begin{table*}[t!]

\centering
\resizebox{\textwidth}{!}{
\begin{tabular}{ccl|cc|cc|cc|cc|cc}
\toprule
\multicolumn{2}{c}{} & \multirow{3}{*}{\textbf{Model}} & \multicolumn{10}{c}{\textbf{Performance by Dialect (CLIPScore) $\uparrow$}} \\
\cmidrule(lr){4-13} 
\multicolumn{2}{c}{} & & \multicolumn{2}{c|}{AAE} & \multicolumn{2}{c|}{BrE} & \multicolumn{2}{c|}{ChE} & \multicolumn{2}{c|}{InE} & \multicolumn{2}{c}{SgE} \\
\multicolumn{2}{c}{} & & \textbf{Dialect} & \textbf{SAE} & \textbf{Dialect} & \textbf{SAE} & \textbf{Dialect} & \textbf{SAE} & \textbf{Dialect} & \textbf{SAE} & \textbf{Dialect} & \textbf{SAE} \\
\midrule
\multirow{16}{*}{\rotatebox[origin=c]{90}{\textbf{Concise Prompts}}}
& \multirow{10}{*}{\rotatebox[origin=c]{90}{\textit{T2I Models}}}
& Stable Diffusion 1.4 & \cellcolor{mygreen!27} 25.46 & \cellcolor{mygreen!36} 27.83 & \cellcolor{mygreen!39} 28.65 & \cellcolor{mygreen!43} 29.79 & \cellcolor{mygreen!25} 24.68 & \cellcolor{mygreen!38} 28.34 & \cellcolor{mygreen!24} 24.34 & \cellcolor{mygreen!41} 29.38 & \cellcolor{mygreen!29} 25.97 & \cellcolor{mygreen!39} 28.64 \\
 &  & Stable Diffusion 1.5 & \cellcolor{mygreen!29} 25.79 & \cellcolor{mygreen!36} 27.95 & \cellcolor{mygreen!39} 28.66 & \cellcolor{mygreen!43} 29.91 & \cellcolor{mygreen!25} 24.7 & \cellcolor{mygreen!38} 28.32 & \cellcolor{mygreen!24} 24.38 & \cellcolor{mygreen!42} 29.44 & \cellcolor{mygreen!29} 25.86 & \cellcolor{mygreen!39} 28.65 \\
 &  & Stable Diffusion 2.1 & \cellcolor{mygreen!28} 25.72 & \cellcolor{mygreen!39} 28.74 & \cellcolor{mygreen!42} 29.67 & \cellcolor{mygreen!47} 30.88 & \cellcolor{mygreen!25} 24.7 & \cellcolor{mygreen!42} 29.44 & \cellcolor{mygreen!27} 25.31 & \cellcolor{mygreen!46} 30.69 & \cellcolor{mygreen!31} 26.46 & \cellcolor{mygreen!42} 29.54 \\
 &  & Stable Diffusion XL & \cellcolor{mygreen!29} 25.81 & \cellcolor{mygreen!39} 28.69 & \cellcolor{mygreen!43} 29.97 & \cellcolor{mygreen!48} 31.21 & \cellcolor{mygreen!27} 25.37 & \cellcolor{mygreen!42} 29.57 & \cellcolor{mygreen!29} 25.85 & \cellcolor{mygreen!48} 31.15 & \cellcolor{mygreen!35} 27.45 & \cellcolor{mygreen!44} 30.23 \\
 &  & Stable Diffusion 3 & \cellcolor{mygreen!27} 25.45 & \cellcolor{mygreen!38} 28.42 & \cellcolor{mygreen!43} 29.89 & \cellcolor{mygreen!47} 30.97 & \cellcolor{mygreen!26} 25.01 & \cellcolor{mygreen!39} 28.74 & \cellcolor{mygreen!26} 25.02 & \cellcolor{mygreen!45} 30.31 & \cellcolor{mygreen!32} 26.8 & \cellcolor{mygreen!42} 29.67 \\
 &  & Stable Diffusion 3.5 Large & \cellcolor{mygreen!28} 25.62 & \cellcolor{mygreen!39} 28.78 & \cellcolor{mygreen!44} 30.25 & \cellcolor{mygreen!49} 31.5 & \cellcolor{mygreen!27} 25.22 & \cellcolor{mygreen!41} 29.42 & \cellcolor{mygreen!28} 25.67 & \cellcolor{mygreen!48} 31.14 & \cellcolor{mygreen!34} 27.18 & \cellcolor{mygreen!44} 30.22 \\
 &  & Stable Diffusion 3.5 Large Turbo & \cellcolor{mygreen!26} 25.1 & \cellcolor{mygreen!38} 28.4 & \cellcolor{mygreen!43} 29.9 & \cellcolor{mygreen!48} 31.16 & \cellcolor{mygreen!26} 24.95 & \cellcolor{mygreen!40} 28.9 & \cellcolor{mygreen!27} 25.3 & \cellcolor{mygreen!46} 30.78 & \cellcolor{mygreen!33} 26.96 & \cellcolor{mygreen!43} 29.82 \\
 &  & Flux.1 [dev] & \cellcolor{mygreen!25} 24.74 & \cellcolor{mygreen!35} 27.54 & \cellcolor{mygreen!38} 28.52 & \cellcolor{mygreen!43} 29.88 & \cellcolor{mygreen!23} 24.21 & \cellcolor{mygreen!36} 27.97 & \cellcolor{mygreen!25} 24.78 & \cellcolor{mygreen!42} 29.48 & \cellcolor{mygreen!27} 25.4 & \cellcolor{mygreen!38} 28.31 \\
 &  & DALL-E Mini & \cellcolor{mygreen!25} 24.77 & \cellcolor{mygreen!37} 28.15 & \cellcolor{mygreen!42} 29.48 & \cellcolor{mygreen!46} 30.65 & \cellcolor{mygreen!21} 23.57 & \cellcolor{mygreen!36} 27.81 & \cellcolor{mygreen!24} 24.4 & \cellcolor{mygreen!42} 29.56 & \cellcolor{mygreen!28} 25.74 & \cellcolor{mygreen!39} 28.6 \\
 &  & DALL-E 2 & \cellcolor{mygreen!24} 24.57 & \cellcolor{mygreen!34} 27.4 & \cellcolor{mygreen!43} 29.86 & \cellcolor{mygreen!45} 30.56 & \cellcolor{mygreen!22} 23.98 & \cellcolor{mygreen!34} 27.3 & \cellcolor{mygreen!23} 24.1 & \cellcolor{mygreen!41} 29.3 & \cellcolor{mygreen!31} 26.44 & \cellcolor{mygreen!38} 28.53 \\
 &  & DALL-E 3 & \cellcolor{mygreen!27} 25.19 & \cellcolor{mygreen!35} 27.51 & \cellcolor{mygreen!40} 28.95 & \cellcolor{mygreen!43} 29.75 & \cellcolor{mygreen!24} 24.57 & \cellcolor{mygreen!37} 28.11 & \cellcolor{mygreen!28} 25.71 & \cellcolor{mygreen!42} 29.66 & \cellcolor{mygreen!30} 26.3 & \cellcolor{mygreen!40} 29.08 \\
 &  & DALL-E 3 w/ Rewrite & \cellcolor{mygreen!26} 24.92 & \cellcolor{mygreen!33} 26.91 & \cellcolor{mygreen!41} 29.41 & \cellcolor{mygreen!44} 30.11 & \cellcolor{mygreen!26} 25.15 & \cellcolor{mygreen!35} 27.57 & \cellcolor{mygreen!32} 26.87 & \cellcolor{mygreen!43} 29.93 & \cellcolor{mygreen!33} 27.12 & \cellcolor{mygreen!38} 28.47 \\
 &  & gpt-image-1 & \cellcolor{mygreen!29} 25.96 & \cellcolor{mygreen!38} 28.33 & \cellcolor{mygreen!47} 30.94 & \cellcolor{mygreen!49} 31.62 & \cellcolor{mygreen!31} 26.51 & \cellcolor{mygreen!42} 29.48 & \cellcolor{mygreen!35} 27.51 & \cellcolor{mygreen!48} 31.21 & \cellcolor{mygreen!38} 28.57 & \cellcolor{mygreen!45} 30.33 \\

\cmidrule(lr){2-13} 
& \multirow{5}{*}{\rotatebox[origin=c]{90}{\textit{T2V Models}}}

& Cosmos-1 & \cellcolor{mygreen!21} 23.49 & \cellcolor{mygreen!27} 25.42 & \cellcolor{mygreen!30} 26.17 & \cellcolor{mygreen!33} 27.16 & \cellcolor{mygreen!18} 22.89 & \cellcolor{mygreen!25} 24.62 & \cellcolor{mygreen!23} 24.18 & \cellcolor{mygreen!33} 27.04 & \cellcolor{mygreen!15} 21.89 & \cellcolor{mygreen!18} 22.91 \\
 &  & Open-Sora & \cellcolor{mygreen!26} 25.02 & \cellcolor{mygreen!34} 27.3 & \cellcolor{mygreen!39} 28.63 & \cellcolor{mygreen!43} 29.73 & \cellcolor{mygreen!24} 24.34 & \cellcolor{mygreen!33} 27.09 & \cellcolor{mygreen!27} 25.35 & \cellcolor{mygreen!41} 29.36 & \cellcolor{mygreen!28} 25.55 & \cellcolor{mygreen!36} 28.01 \\
 &  & VideoCrafter-2 & \cellcolor{mygreen!29} 25.88 & \cellcolor{mygreen!39} 28.83 & \cellcolor{mygreen!41} 29.41 & \cellcolor{mygreen!46} 30.69 & \cellcolor{mygreen!26} 25.04 & \cellcolor{mygreen!40} 29.04 & \cellcolor{mygreen!29} 25.88 & \cellcolor{mygreen!45} 30.56 & \cellcolor{mygreen!33} 27 & \cellcolor{mygreen!42} 29.69 \\
 &  & CogVideoX & \cellcolor{mygreen!17} 22.62 & \cellcolor{mygreen!28} 25.71 & \cellcolor{mygreen!23} 24.14 & \cellcolor{mygreen!29} 25.84 & \cellcolor{mygreen!15} 22.03 & \cellcolor{mygreen!24} 24.61 & \cellcolor{mygreen!19} 22.95 & \cellcolor{mygreen!34} 27.4 & \cellcolor{mygreen!8} 19.99 & \cellcolor{mygreen!16} 22.18 \\
 &  & Wan 2.1 & \cellcolor{mygreen!17} 22.37 & \cellcolor{mygreen!28} 25.49 & \cellcolor{mygreen!27} 25.45 & \cellcolor{mygreen!37} 28.27 & \cellcolor{mygreen!16} 22.14 & \cellcolor{mygreen!24} 24.55 & \cellcolor{mygreen!17} 22.55 & \cellcolor{mygreen!35} 27.57 & \cellcolor{mygreen!18} 22.75 & \cellcolor{mygreen!32} 26.85 \\

\midrule
\multirow{16}{*}{\rotatebox[origin=c]{90}{\textbf{Detailed Prompts}}}
& \multirow{10}{*}{\rotatebox[origin=c]{90}{\textit{T2I Models}}}

& Stable Diffusion 1.4 & \cellcolor{mygreen!36} 27.98 & \cellcolor{mygreen!39} 28.84 & \cellcolor{mygreen!42} 29.59 & \cellcolor{mygreen!44} 30.23 & \cellcolor{mygreen!39} 28.61 & \cellcolor{mygreen!44} 30.1 & \cellcolor{mygreen!32} 26.79 & \cellcolor{mygreen!43} 29.83 & \cellcolor{mygreen!37} 28.12 & \cellcolor{mygreen!43} 29.78 \\
 &  & Stable Diffusion 1.5 & \cellcolor{mygreen!37} 28.08 & \cellcolor{mygreen!40} 28.99 & \cellcolor{mygreen!42} 29.54 & \cellcolor{mygreen!45} 30.29 & \cellcolor{mygreen!38} 28.52 & \cellcolor{mygreen!44} 30.18 & \cellcolor{mygreen!32} 26.87 & \cellcolor{mygreen!43} 29.94 & \cellcolor{mygreen!36} 27.98 & \cellcolor{mygreen!43} 29.82 \\
 &  & Stable Diffusion 2.1 & \cellcolor{mygreen!38} 28.57 & \cellcolor{mygreen!43} 29.9 & \cellcolor{mygreen!46} 30.83 & \cellcolor{mygreen!49} 31.69 & \cellcolor{mygreen!42} 29.68 & \cellcolor{mygreen!49} 31.51 & \cellcolor{mygreen!37} 28.24 & \cellcolor{mygreen!49} 31.47 & \cellcolor{mygreen!42} 29.54 & \cellcolor{mygreen!48} 31.32 \\
 &  & Stable Diffusion XL & \cellcolor{mygreen!38} 28.46 & \cellcolor{mygreen!42} 29.68 & \cellcolor{mygreen!45} 30.49 & \cellcolor{mygreen!48} 31.33 & \cellcolor{mygreen!40} 29.12 & \cellcolor{mygreen!48} 31.14 & \cellcolor{mygreen!36} 27.95 & \cellcolor{mygreen!47} 30.96 & \cellcolor{mygreen!39} 28.65 & \cellcolor{mygreen!45} 30.52 \\
 &  & Stable Diffusion 3 & \cellcolor{mygreen!39} 28.69 & \cellcolor{mygreen!43} 29.82 & \cellcolor{mygreen!46} 30.76 & \cellcolor{mygreen!49} 31.59 & \cellcolor{mygreen!40} 29.13 & \cellcolor{mygreen!48} 31.15 & \cellcolor{mygreen!36} 27.82 & \cellcolor{mygreen!47} 31 & \cellcolor{mygreen!40} 29.13 & \cellcolor{mygreen!47} 31.04 \\
 &  & Stable Diffusion 3.5 Large & \cellcolor{mygreen!39} 28.86 & \cellcolor{mygreen!44} 30.01 & \cellcolor{mygreen!47} 31.02 & \cellcolor{mygreen!50} 31.84 & \cellcolor{mygreen!42} 29.48 & \cellcolor{mygreen!49} 31.64 & \cellcolor{mygreen!37} 28.04 & \cellcolor{mygreen!49} 31.61 & \cellcolor{mygreen!41} 29.29 & \cellcolor{mygreen!48} 31.19 \\
 &  & Stable Diffusion 3.5 Large Turbo & \cellcolor{mygreen!39} 28.61 & \cellcolor{mygreen!42} 29.58 & \cellcolor{mygreen!46} 30.67 & \cellcolor{mygreen!49} 31.6 & \cellcolor{mygreen!40} 29.09 & \cellcolor{mygreen!48} 31.23 & \cellcolor{mygreen!36} 27.8 & \cellcolor{mygreen!48} 31.18 & \cellcolor{mygreen!39} 28.76 & \cellcolor{mygreen!46} 30.78 \\
 &  & Flux.1 [dev] & \cellcolor{mygreen!36} 27.97 & \cellcolor{mygreen!39} 28.69 & \cellcolor{mygreen!42} 29.54 & \cellcolor{mygreen!45} 30.37 & \cellcolor{mygreen!37} 28.17 & \cellcolor{mygreen!44} 30.17 & \cellcolor{mygreen!33} 27.15 & \cellcolor{mygreen!44} 30.01 & \cellcolor{mygreen!35} 27.72 & \cellcolor{mygreen!42} 29.46 \\
 &  & DALL-E Mini & \cellcolor{mygreen!34} 27.26 & \cellcolor{mygreen!41} 29.18 & \cellcolor{mygreen!43} 29.84 & \cellcolor{mygreen!45} 30.56 & \cellcolor{mygreen!36} 27.75 & \cellcolor{mygreen!44} 30.23 & \cellcolor{mygreen!32} 26.71 & \cellcolor{mygreen!43} 29.93 & \cellcolor{mygreen!34} 27.42 & \cellcolor{mygreen!42} 29.59 \\
 &  & DALL-E 2 & \cellcolor{mygreen!35} 27.66 & \cellcolor{mygreen!40} 29.02 & \cellcolor{mygreen!45} 30.5 & \cellcolor{mygreen!48} 31.3 & \cellcolor{mygreen!38} 28.57 & \cellcolor{mygreen!45} 30.54 & \cellcolor{mygreen!31} 26.48 & \cellcolor{mygreen!44} 30.17 & \cellcolor{mygreen!39} 28.69 & \cellcolor{mygreen!43} 29.88 \\
 &  & DALL-E 3 & \cellcolor{mygreen!36} 27.79 & \cellcolor{mygreen!38} 28.3 & \cellcolor{mygreen!42} 29.55 & \cellcolor{mygreen!44} 30.21 & \cellcolor{mygreen!38} 28.52 & \cellcolor{mygreen!44} 30.04 & \cellcolor{mygreen!35} 27.48 & \cellcolor{mygreen!43} 29.83 & \cellcolor{mygreen!39} 28.67 & \cellcolor{mygreen!44} 30.03 \\
 &  & DALL-E 3 w/ Rewrite & \cellcolor{mygreen!35} 27.71 & \cellcolor{mygreen!37} 28.23 & \cellcolor{mygreen!43} 29.75 & \cellcolor{mygreen!45} 30.46 & \cellcolor{mygreen!40} 28.88 & \cellcolor{mygreen!43} 29.85 & \cellcolor{mygreen!38} 28.57 & \cellcolor{mygreen!43} 29.98 & \cellcolor{mygreen!39} 28.61 & \cellcolor{mygreen!41} 29.42 \\
 &  & gpt-image-1 & \cellcolor{mygreen!39} 28.65 & \cellcolor{mygreen!42} 29.45 & \cellcolor{mygreen!48} 31.2 & \cellcolor{mygreen!49} 31.6 & \cellcolor{mygreen!43} 29.98 & \cellcolor{mygreen!48} 31.29 & \cellcolor{mygreen!41} 29.41 & \cellcolor{mygreen!46} 30.81 & \cellcolor{mygreen!44} 30.18 & \cellcolor{mygreen!48} 31.27 \\

\cmidrule(lr){2-13} 
& \multirow{5}{*}{\rotatebox[origin=c]{90}{\textit{T2V Models}}}

& Cosmos-1 & \cellcolor{mygreen!19} 23.07 & \cellcolor{mygreen!22} 23.79 & \cellcolor{mygreen!29} 25.98 & \cellcolor{mygreen!32} 26.67 & \cellcolor{mygreen!23} 24.19 & \cellcolor{mygreen!26} 24.94 & \cellcolor{mygreen!20} 23.35 & \cellcolor{mygreen!27} 25.29 & \cellcolor{mygreen!8} 19.99 & \cellcolor{mygreen!12} 21.09 \\
 &  & Open-Sora & \cellcolor{mygreen!34} 27.4 & \cellcolor{mygreen!38} 28.36 & \cellcolor{mygreen!42} 29.5 & \cellcolor{mygreen!43} 29.93 & \cellcolor{mygreen!37} 28.07 & \cellcolor{mygreen!42} 29.46 & \cellcolor{mygreen!35} 27.64 & \cellcolor{mygreen!44} 30.04 & \cellcolor{mygreen!35} 27.64 & \cellcolor{mygreen!41} 29.19 \\
 &  & VideoCrafter-2 & \cellcolor{mygreen!38} 28.4 & \cellcolor{mygreen!43} 29.76 & \cellcolor{mygreen!44} 30.24 & \cellcolor{mygreen!47} 30.98 & \cellcolor{mygreen!40} 28.95 & \cellcolor{mygreen!47} 31 & \cellcolor{mygreen!36} 27.83 & \cellcolor{mygreen!47} 30.88 & \cellcolor{mygreen!39} 28.74 & \cellcolor{mygreen!46} 30.61 \\
 &  & CogVideoX & \cellcolor{mygreen!13} 21.42 & \cellcolor{mygreen!17} 22.55 & \cellcolor{mygreen!24} 24.38 & \cellcolor{mygreen!28} 25.74 & \cellcolor{mygreen!18} 22.89 & \cellcolor{mygreen!24} 24.6 & \cellcolor{mygreen!17} 22.37 & \cellcolor{mygreen!28} 25.51 & 17.67 & \cellcolor{mygreen!8} 19.82 \\
 &  & Wan 2.1 & \cellcolor{mygreen!29} 25.85 & \cellcolor{mygreen!36} 27.89 & \cellcolor{mygreen!36} 27.96 & \cellcolor{mygreen!41} 29.2 & \cellcolor{mygreen!30} 26.05 & \cellcolor{mygreen!39} 28.83 & \cellcolor{mygreen!27} 25.25 & \cellcolor{mygreen!40} 28.92 & \cellcolor{mygreen!27} 25.45 & \cellcolor{mygreen!36} 27.98 \\
 
\bottomrule
\end{tabular}
}

\caption{\textbf{Complete \Dataset{} Benchmark Performance Breakdown} by dialect for all text-to-image and text-to-video generative models. All performance scores are measured using CLIPScore~\citep{Hessel2021CLIPScoreAR}, higher score is better. Results complement \Cref{tab:main-benchmark-results} in the main paper.}
\label{tab:benchmark-results-clip} 
\end{table*}

Due to space constraints in the main paper, we report per-dialect breakdowns multimodal generative model performances on the \Dataset{} benchmark in \Cref{tab:mitigation-methods-detail}, \Cref{tab:benchmark-results-vqa}, and \Cref{tab:benchmark-results-clip}.

As described in \Cref{subsec:evaluation_metrics}, the scoring functions are based on reference-free image-text alignment metrics, including VQAScore and CLIPScore. We denote the subset of \Dataset{} prompts corresponding to a given dialect as $\mathcal{P}$, which consists of multiple SAE Prompt / Dialect Prompt pairs $p = (p^s, p^d)$.
For each individual text prompt $p^s$ or $p^d$, we generate $n$ images under different random seeds for text-to-image generative models, or uniformly sample $n$ frames for text-to-video generative models.

Accordingly, for each SAE Prompt / Dialect Prompt pair $p = (p^s, p^d) \in \mathcal{P}$, we compute its SAE and Dialect performances using \Cref{eq:performance}.
More concretely, $SAE(p,\ \mathcal{G})$ in \Cref{eq:performance} denotes the average VQAScore (as reported in \Cref{tab:mitigation-methods-detail} and \Cref{tab:benchmark-results-vqa}) or CLIPScore (in \Cref{tab:benchmark-results-clip}) computed over the $n$ images generated from the SAE prompt $p^s$.
Similarly, $Dialect(p,\ \mathcal{G})$ in \Cref{eq:performance} is computed using the same evaluation pipeline, but with the corresponding dialect prompt $p^d$ from the same pair.
Each value of $SAE(p,\ \mathcal{G})$ and $Dialect(p,\ \mathcal{G})$ is reported as \textbf{SAE} and \textbf{Dialect}, respectively, in the tables.

\section{Grammatical vs. Lexical Robustness in Multimodal Models}
\label{subsec:grammatical}
To establish the rationale for our study's focus on lexical variations, we begin with an observation about multimodal generative models. These models often exhibit a notable insensitivity to grammatical or syntactic structure, a tendency that likely arises from the bag-of-words nature of their CLIP-style encoders.
This architectural trait means that variations in sentence construction, such as word order or verb tenses, tend to have a minimal effect on the final output. \Cref{tab:grammatial_var_def}, adapted from Multi-VALUE~\citep{Ziems2023MultiVALUEAF}, showcases several examples of these grammatical variations.

To formally quantify this observation, we conducted a small-scale experiment with three representative models in the African American English evaluation setting. We used the Multi-VALUE~\citep{Ziems2023MultiVALUEAF} translation system to apply grammatical variations to 300 SAE prompts from \Dataset{} and evaluated their generation quality using VQAScore.

The results, presented in \Cref{tab:grammatical_var_res}, provide strong quantitative evidence supporting our initial analysis. While \textbf{lexical feature variations cause significant performance drops} for existing text-to-image generative models, \textbf{grammatical variations do not incur significant performance drops.} This clear distinction validates our decision to focus on the more impactful lexical variations throughout this work.
\begin{table*}[htbp]
\centering
\begin{tabular}{@{}lll@{}}
\toprule
\textbf{Grammatical Variation Type} & \textbf{SAE Prompt} & \textbf{AAE Dialect Prompt} \\ \midrule
Clause Structure & A chair \textbf{\textcolor{my_blue}{that}} can be folded & A chair can be folded \\
Negative Concord & There \textbf{\textcolor{my_blue}{is}} no food on the table & There \textbf{\textcolor{my_purple}{ain't}} no food on the table \\
Word Order & A \textbf{\textcolor{my_blue}{big and fresh}} fish & A fish \textbf{\textcolor{my_purple}{big and fresh}} \\
Verb Morphology & Mom \textbf{\textcolor{my_blue}{brought}} rice to me & Mom \textbf{\textcolor{my_purple}{brin}} rice give me \\ \bottomrule
\end{tabular}

\caption{\textbf{Examples of Grammatical Dialect Variations} between Standard American English (SAE) sentences and African American English (AAE) dialect sentences. The \textbf{\textcolor{my_blue}{blue}} texts highlight unique features in SAE while the \textbf{\textcolor{my_purple}{purple}} texts (if applicable) highlight corresponding features in AAE.}
\label{tab:grammatial_var_def}
\end{table*}
\begin{table*}[htbp]
\centering
\resizebox{\textwidth}{!}{%
\begin{tabular}{@{}lcccc@{}}
\toprule
\multirow{2}{*}{\textbf{Model}} & \multirow{2}{*}{\textbf{SAE Performance (\%)}} & \multicolumn{3}{c}{\textbf{Performance under Dialectal Variations (\%)}} \\ 
\cmidrule(lr){3-5}
& & \textbf{Grammatical} & \textbf{Lexical} & \textbf{Grammatical + Lexical} \\ 
\midrule
DALL-E Mini & 75.63 & 74.72 (-1.20) & 51.92 (-31.35) & 51.26 (-32.22) \\
FLUX.1 dev & 82.94 & 82.40 (-0.65) & 61.88 (-25.39) & 61.02 (-26.43) \\
Stable Diffusion 3.5 Large & 85.18 & 83.91 (-1.49) & 65.37 (-23.26) & 63.80 (-25.10) \\ \bottomrule
\end{tabular}%
}

\caption{\textbf{Quantitative Effects of Grammatical and Lexical Variations on Multimodal Generation}, measured in VQAScore. We evaluate three text-to-image generative models under the following dialectal variation types: Grammatical, Lexical, and Grammatical + Lexical. Values in parentheses indicate the percentage performance drop in VQAScore compared to baseline SAE performance.}
\label{tab:grammatical_var_res}
\end{table*}

\begin{table*}[htbp]
\centering
\small
\resizebox{\textwidth}{!}{%
\begin{tabular}{@{}lcccc@{}}
\toprule
\multirow{2}{*}{\textbf{Model}} & \multirow{2}{*}{\textbf{SAE Prompt Performance (\%)}} & \multicolumn{2}{c}{\textbf{Dialectal Prompt Performance (\%)}} \\ 
\cmidrule(lr){3-4}
& & \textbf{Single Dialect Lexeme} & \textbf{Multi Dialect Lexeme} \\ 
\midrule
DALL·E Mini & 72.39 & 48.65 (-32.79) & 32.47 (-55.15) \\
FLUX.1 dev & 83.04 & 54.81 (-34.00) & 38.90 (-46.26) \\
Stable Diffusion 3.5 Large & 88.73 & 56.21 (-36.65) & 37.62 (-57.60) \\ 
\bottomrule
\end{tabular}%
}
\caption{\textbf{Quantitative Effects of Single vs. Multiple Dialectal Lexeme Variations on Multimodal Generation}, measured in VQAScore. We evaluate three text-to-image generative models on 100 newly annotated Indian English SAE–dialect prompt pairs. The Single Dialect Lexeme setting replaces one SAE word with its dialectal equivalent, while the Multi Dialect Lexeme setting replaces 2–3 words. Values in parentheses indicate the percentage performance drop relative to SAE prompts.}
\label{tab:multi_word_var_res}
\end{table*}

\section{Compositional Impact of Multiple Dialect Lexemes on Multimodal Generative Models}
\label{subsec:multi-word}

In this section we share results of a quantitative experiment on 100 newly annotated SAE-dialect prompt pairs from the Indian English setting. Each prompt contains 2-3 SAE lexemes with dialect word equivalents. In the Single Dialect Word setting, we replace one of them with its dialect word equivalent, and in the Multi Dialect Word setting we replace 2-3 words. Finally, we evaluate each model’s generation quality with the VQAScore metric.

The results are shown in ~\Cref{tab:multi_word_var_res}. In line with our expectations, models experience more significant performance degradation when more dialectal lexemes are introduced in the prompt. These results highlight the challenging nature of the dialect robustness problem in multimodal generation. It further hints at the compositional impact of the issue, where dialect understanding errors become more severe when more dialectal lexicons are involved. We look forward to further exploring this topic in our future works.

\section{Text Encoder Training Data Analysis}
\label{app:text_encoder_data_analysis}

To diagnose the dialect-wise performance gaps observed in \Cref{tab:main-benchmark-results}, we analyze whether dialect lexemes are underrepresented in the training corpora of text encoders used by the evaluated models. Stable Diffusion 1.5 and SDXL both use the CLIP ViT-L/14~\citep{radford2021learning} text encoder, whose full training corpus is not publicly released. SDXL also uses a second OpenCLIP ViT-bigG/14~\citep{ilharco2021openclip} encoder, trained on LAION data~\citep{schuhmann2022laion}. Wan 2.1 uses umt5-xxl~\citep{chung2023unimax}, trained on mC4~\citep{chung2023unimax}. Given limited public availability of text encoder training datasets, we use LAION-2B~\citep{schuhmann2022laion} and Multilingual C4~\citep{chung2023unimax} as representative training data for evaluated model. While this does not fully reconstruct all encoder pretraining data, it provides a starting point for analyzing dialect data prevalence in training datasets.

We define two complementary metrics corresponding to the two lexeme types in \Dataset{}. For \emph{non-polysemous} lexemes, a dialect word either appears in the corpus or it does not, so we can directly compare raw frequencies against SAE counterparts. For \emph{polysemous} lexemes, however, a dialect word may appear frequently but be used overwhelmingly in its SAE sense rather than its dialect sense; this distinction requires manual inspection. We therefore define separate metrics for each case.

\paragraph{Pairwise Dialect Share (PDS).}
For non-polysemous lexemes, let each SAE/dialect pair be $\ell = (w^s, w^d)$, and let $f_C(w)$ denote the frequency of $w$ in corpus $C$. We define:
\begin{equation}
    PDS(\ell,\ C) = \frac{f_C(w^d)}{f_C(w^s) + f_C(w^d)}.
    \label{eq:pds_pair}
\end{equation}
The dialect-level PDS averages over all non-polysemous pairs $\mathcal{L}_d^{\mathrm{np}}$ for dialect $d$. A balanced corpus would yield $PDS = 0.50$.

\paragraph{Dialect Sense Rate (DSR).}
For polysemous lexemes, we retrieve corpus contexts containing each dialect lexeme $\ell$ and \emph{manually label} each usage as expressing either the dialect meaning or the SAE meaning. Let $N_D(\ell, C)$ and $N_S(\ell, C)$ denote the counts of dialect-sense and SAE-sense usages, respectively. We define:
\begin{equation}
    DSR(\ell,\ C) = \frac{N_D(\ell,\ C)}{N_D(\ell,\ C) + N_S(\ell,\ C)}.
    \label{eq:dsr_lexeme}
\end{equation}
The dialect-level DSR averages over all polysemous lexemes $\mathcal{L}_d^{\mathrm{poly}}$ for dialect $d$. A balanced corpus would yield $DSR = 0.50$.

\begin{table}[t]
\centering
\small
\setlength{\tabcolsep}{6pt}
\begin{tabular}{l|cc|cc|cc|cc|cc}
\toprule
\multirow{2}{*}{\textbf{Training Data}} & \multicolumn{2}{c|}{AAE} & \multicolumn{2}{c|}{BrE} & \multicolumn{2}{c|}{ChE} & \multicolumn{2}{c|}{InE} & \multicolumn{2}{c}{SgE} \\
& \textbf{PDS} & \textbf{DSR} & \textbf{PDS} & \textbf{DSR} & \textbf{PDS} & \textbf{DSR} & \textbf{PDS} & \textbf{DSR} & \textbf{PDS} & \textbf{DSR} \\
\midrule
Balanced Data & 0.50 & 0.50 & 0.50 & 0.50 & 0.50 & 0.50 & 0.50 & 0.50 & 0.50 & 0.50 \\
\midrule
LAION-2B & 0.09 & 0.03 & 0.37 & 0.26 & 0.01 & 0.00 & 0.03 & 0.00 & 0.07 & 0.00 \\
mC4      & 0.03 & 0.01 & 0.29 & 0.15 & 0.01 & 0.00 & 0.02 & 0.00 & 0.05 & 0.00 \\

\bottomrule
\end{tabular}
\caption{Text encoder training data analysis. \textbf{PDS} (Pairwise Dialect Share) measures the relative frequency of non-polysemous dialect lexemes compared to their SAE counterparts, and \textbf{DSR} (Dialect Sense Rate) measures how often polysemous dialect lexemes appear in their dialect sense rather than their SAE sense. Both metrics are averaged per dialect; a perfectly balanced corpus would score $0.50$ on both. Results are reported on LAION-2B and mC4 as representative public training datasets.}
\label{tab:appendix_text_encoder_analysis}
\end{table}

\paragraph{Analysis.}
As shown in \Cref{tab:appendix_text_encoder_analysis}, dialect lexeme exposure is severely imbalanced across all dialects and both corpora. ChE and InE exhibit the most extreme underrepresentation: PDS values of $0.01$--$0.03$ indicate that dialect lexemes constitute less than 3\% of all occurrences of their referred concepts. Their DSR values are exactly $0.00$ in both corpora, meaning that even when polysemous ChE and InE lexemes do appear, they are \emph{never} used in the dialect sense. This directly corresponds to the performance patterns in \Cref{tab:main-benchmark-results}, where ChE and InE consistently incur the largest performance drops across nearly all models (e.g., 34--55\% VQAScore drops for concise prompts).

AAE and SgE occupy a slightly better position, with PDS values of $0.03$--$0.09$ and DSR near zero ($\leq 0.03$), consistent with their moderate but still substantial performance drops in \Cref{tab:main-benchmark-results} (roughly 20--33\% for most T2I models). In contrast, BrE stands out with markedly higher exposure: PDS of $0.37$ (LAION-2B) and $0.29$ (mC4), and DSR of $0.26$ and $0.15$. This aligns with BrE showing the smallest dialect-induced drops in \Cref{tab:main-benchmark-results}, often below 10\%.

Finally, mC4 shows uniformly lower PDS and DSR than LAION-2B across all dialects. This is notable because Wan 2.1, whose umt5-xxl encoder is pretrained on mC4, exhibits the most severe overall performance drops among all evaluated models (47.33\% VQAScore drop on concise prompts in \Cref{tab:main-benchmark-results}), suggesting that its encoder's lower dialect data exposure may have contributed to its poor dialect robustness.

\section{Implementation Details}
\label{subsec:implementation_details}

\paragraph{Data Preparation}
We first split the \Dataset{} dataset into training, validation, and test sets in a ratio of 80\%, 10\%, and 10\%, respectively. These training and validation splits of \Dataset{} are used to compute the Dialect Learning loss and the Polysemy Control loss. For KL Regularization loss, we randomly sample 1,024 and 256 image-caption pairs from the MSCOCO validation set~\citep{lin2014microsoft} for use in training and validation, respectively. The target text encoder is evaluated on the validation set at the end of each epoch, and the checkpoint with the lowest validation loss is selected and saved for final evaluation. We then evaluate SAE polysemy and per-dialect performance using the test split of \Dataset{}, and assess SAE MSCOCO performance on 50 randomly sampled captions from the MSCOCO validation set.

\paragraph{Training}
We employ the pretrained text encoder and fine-tune it for 30 epochs using the AdamW optimizer with an initial learning rate of $1\times10^{-4}$, $\beta_1 = 0.9$, $\beta_2 = 0.999$, and $\epsilon = 1\times10^{-8}$. A cosine annealing learning rate scheduler is applied across the 30 training epochs.
The batch size, \textit{i.e.}, $N$ in \Cref{eq:loss-dialect} and \Cref{eq:loss-polysemy}, is set to 32, and the number of image-caption pairs used for KL regularization, \textit{i.e.}, $M$ in \Cref{eq:loss-kl}, is set to 1,024. Training is completed in less than one hour on a single NVIDIA RTX A6000 GPU. In the case of SDXL, which includes both Base and Refiner encoders, the number of pairs $M$ for the Refiner encoder is set to 512 due to its larger size, and training takes approximately one hour using four NVIDIA RTX A6000 GPUs, with all other configurations kept the same as in the Stable Diffusion 1.5 and SDXL Base encoder settings.

\paragraph{About T2Video Models}
Video‐generation models incur substantially higher computational cost than their image counterparts. Since our primary goal is to assess the models’ ability to interpret and render textual prompts, we generate only a small, fixed number of frames per video. This strategy is justified by two observations: (i) the first few frames typically suffice to judge prompt fidelity, and (ii) our prompts do not exhibit extensive motion, so long sequences offer diminishing returns.

All models were obtained by cloning their official repositories and following the authors’ installation instructions. We uniformly reduced frame counts when possible. In some cases, spatial resolution was also reduced to facilitate efficient evaluation—see \Cref{tab:text2video_params} for the precise settings. 

Average time per video was measured on a single NVIDIA RTX A6000 GPU; the Wan2.1‐T2V‐14B model, which does not fit in single‐GPU memory, was benchmarked using six A6000 GPUs under Fully Sharded Data Parallel (FSDP) supported by the repository under the xdit framework.

All models except Wan2.1 fit under a single A6000 GPU and use approximately 20-30 GB of VRAM max. Wan2.1 takes at least 3 GPUs, taking an approximate memory usage of ~100GB of combined VRAM.

\section{Model Details}
\label{subsec:models_details}
We provide detailed information on the multimodal generative models and key experimental settings used in our benchmark.

\Cref{tab:model_specs_reordered} lists the comprehensive specifications for all models evaluated in our work, including both text-to-image and text-to-video models. For each model, we provide details such as its creator organization, initial release date, hosting platform, availability type (\textit{e.g.}, open source, proprietary), and model size.

\Cref{tab:text2video_params} describes in detail the key generation parameters used for the text-to-video models. This includes the specific resolution, number of frames, and inference steps used for each model. Furthermore, we specify the average time required to generate a single video and the total time needed to generate our full video dataset to aid in understanding the reproducibility and computational cost of our experiments.
\begin{table*}[htbp]

\centering
\resizebox{\textwidth}{!}{
\begin{tabular}{@{}lllllll@{}}
\toprule
\textbf{Model Name} & \textbf{Model Type} & \textbf{Created by} & \textbf{Release Date} & \textbf{Hosted by} & \textbf{Availability Type} & \textbf{Model Size} \\
\midrule
Stable Diffusion 1.4 & Text to Image & CompVis & 8/22/2022 & Hugging Face & Open Source & 1B \\
Stable Diffusion 1.5 & Text to Image & Runway ML & 10/20/2022 & Hugging Face & Open Weights & 1.3 B \\
Stable Diffusion 2.1 & Text to Image & Stability AI & 12/7/2022 & Hugging Face & Open Weights & 1.3 B \\
Stable Diffusion XL & Text to Image & Stability AI & 7/26/2023 & Hugging Face & Open Weights & 6.6 B \\
Stable Diffusion 3 Medium & Text to Image & Stability AI & 6/12/2024 & Hugging Face & Open Weights & 2 B \\
Stable Diffusion 3.5 Large & Text to Image & Stability AI & 10/22/2024 & Hugging Face & Open Weights & 8.1 B \\
Stable Diffusion 3.5 Large Turbo & Text to Image & Stability AI & 10/22/2024 & Hugging Face & Open Weights & 8.1 B \\
Flux.1 [dev] & Text to Image & Black Forest Labs & 4/2/2024 & Hugging Face & Open Weights & 12B \\
DALL-E Mini & Text to Image & Boris Dayma et al. & 7/25/2022 & Github & Open Weights & 0.4 B \\
DALL-E 2 & Text to Image & OpenAI & 9/28/2022 & OpenAI & Proprietary & N/A \\
DALL-E 3 & Text to Image & OpenAI & 8/20/2023 & OpenAI & Proprietary & N/A \\
gpt-image-1 & Text to Image & OpenAI & 4/23/2025 & OpenAI & Proprietary & N/A \\
VideoCrafter-2 & Text to Video & Tencent & 1/26/2024 & Hugging Face & Open Weights & 1.4 B \\
Open-Sora & Text to Video & HPC-AI Tech & 6/17/2024 & Hugging Face & Open Weights & 1.2 B \\
CogVideoX & Text to Video & THUDM Lab & 8/27/2024 & Hugging Face & Open Weights & 5 B \\
Cosmos-1 & Text to Video & Nvidia & 1/6/2025 & Hugging Face & Open Weights & 7 B \\
Wan 2.1 & Text to Video & Alibaba & 2/22/2025 & Hugging Face & Open Weights & 14 B \\
\bottomrule
\end{tabular}%
}
\caption{\textbf{Detailed Model Specifications} for all multimodal generative models (text-to-image and text-to-video generative models) benchmarked in this work. For reference and reproducibility, we include model name, model type, creator organization, initial release date, hosting platform, availability type, and model size.}
\label{tab:model_specs_reordered}
\end{table*}

\begin{table*}[htbp]

  \centering
  \resizebox{\textwidth}{!}{%
    \begin{tabular}{lccccr}
      \toprule
      \textbf{Model Version} & \textbf{Resolution} & \textbf{Frames} & \textbf{Steps} & \textbf{Time / Video (min)} & \textbf{Time / Dataset (h)} \\
      \midrule
      VideoCrafter2               & $512\times512$   & 16  & 50 &  5.0  &  342.5 \\
      OpenSora‐STDiT‐v3                         & $405\times720$   & 51  & 30 &  8.3  &  570.8 \\
      CogVideoX‐5b                              & $720\times480$   & 10  & 10 &  6.1  &  416.7 \\
      Cosmos‐1.0‐Diffusion‐7B‐Text2World         & $704\times1280$  & 121 & 35 & 26.5  & 1815.3 \\
      Wan2.1‐T2V‐14B                            & $832\times480$   & 10  & 12 &  4.8  &  329.4 \\
      \bottomrule
    \end{tabular}%
  }
  \footnotesize\textit{Note: The dataset‐scale timing for Wan2.1-T2V-14B was measured using 6 A6000 GPUs using xdit FSDP.}
  \caption{\textbf{Key Generation Parameters for Text-to-Video Generative Models}. For reproducibility and computational cost estimation, we list GPU runtime per video in minutes and GPU runtime for the full video dataset (both concise and detailed = 4110 videos) in hours. All computational costs are estimated for NVIDIA-A6000 GPUs with 48 GB Memory.}
  \label{tab:text2video_params}
\end{table*}

\section{Dataset Details}
\label{subsec:dataset_details}

The final \Dataset{} Dataset contains a total of 4632 prompts, which include 2100 non-SAE dialect prompts, 2100 SAE prompts, and 432 polysemous SAE prompts. The entire dataset is split into three subsets: training, validation, and test. The data split ratio is train : validation : test = 8 : 1 : 1. All benchmarking experiments are performed on the entire dataset, while for mitigation experiments, models are trained on the \Dataset{} training set while evaluated on the validation set.

\section{Human Annotation Details}
\label{subsec:human_eval_details}

\begin{figure*}[htbp]
\centering
    \includegraphics[width=1.0\linewidth]{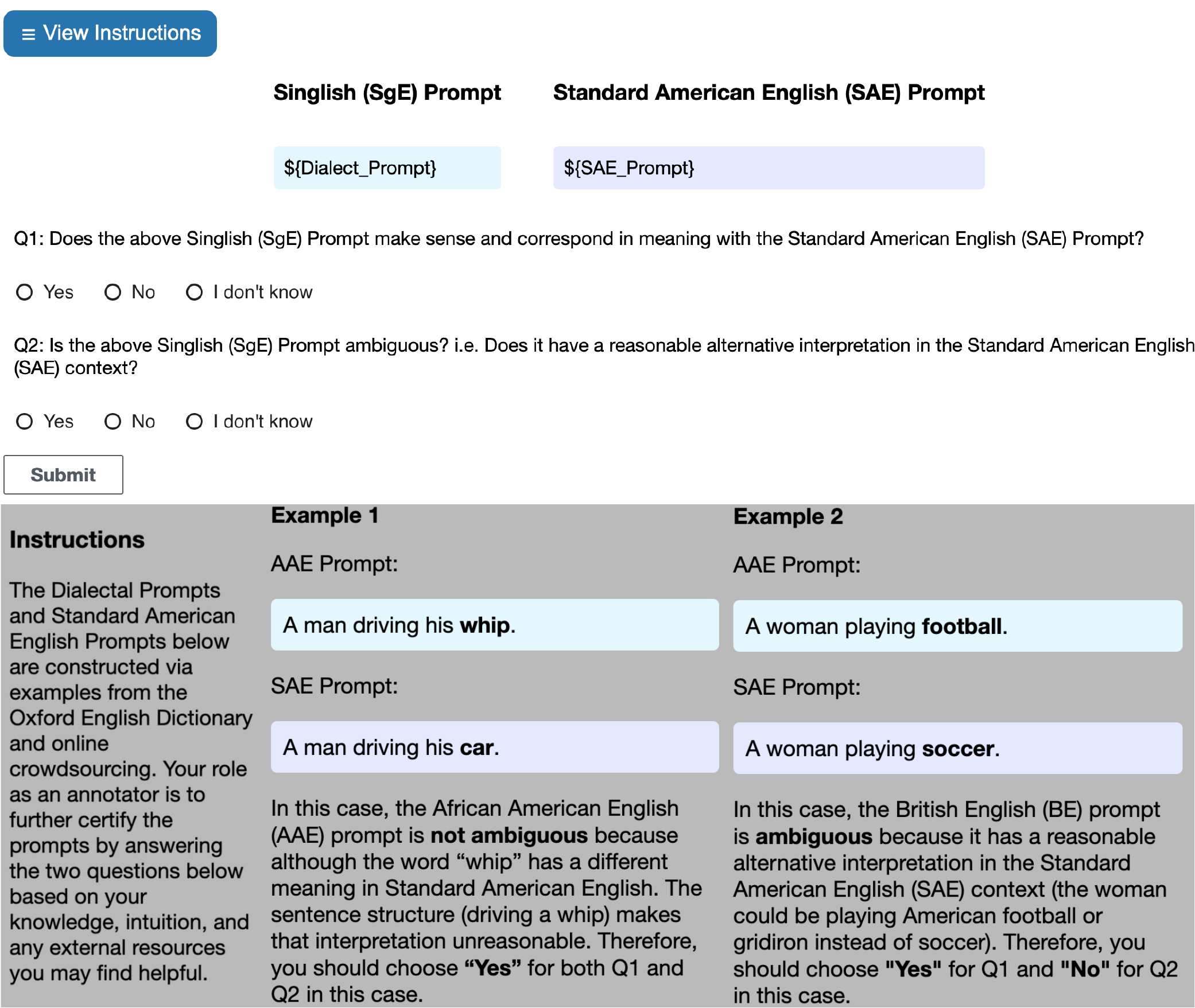}
    \caption{
    \textbf{The Amazon Mechanical Turk Data Annotation Interface} for dialect speaker human filtering of generated prompts (prompt generation details in \Cref{sec:dataset}). Human annotators may use the "View Instructions" button to collapse / re-open detailed annotation instructions at any time. The annotation interface places no maximum time limit on each annotation question. Human annotators are allowed to return to previously annotated questions and update their answers at any time.
    }
    \label{fig:mturk}
\end{figure*}
\begin{figure*}[t!]
\centering
    \includegraphics[width=1.0\linewidth]{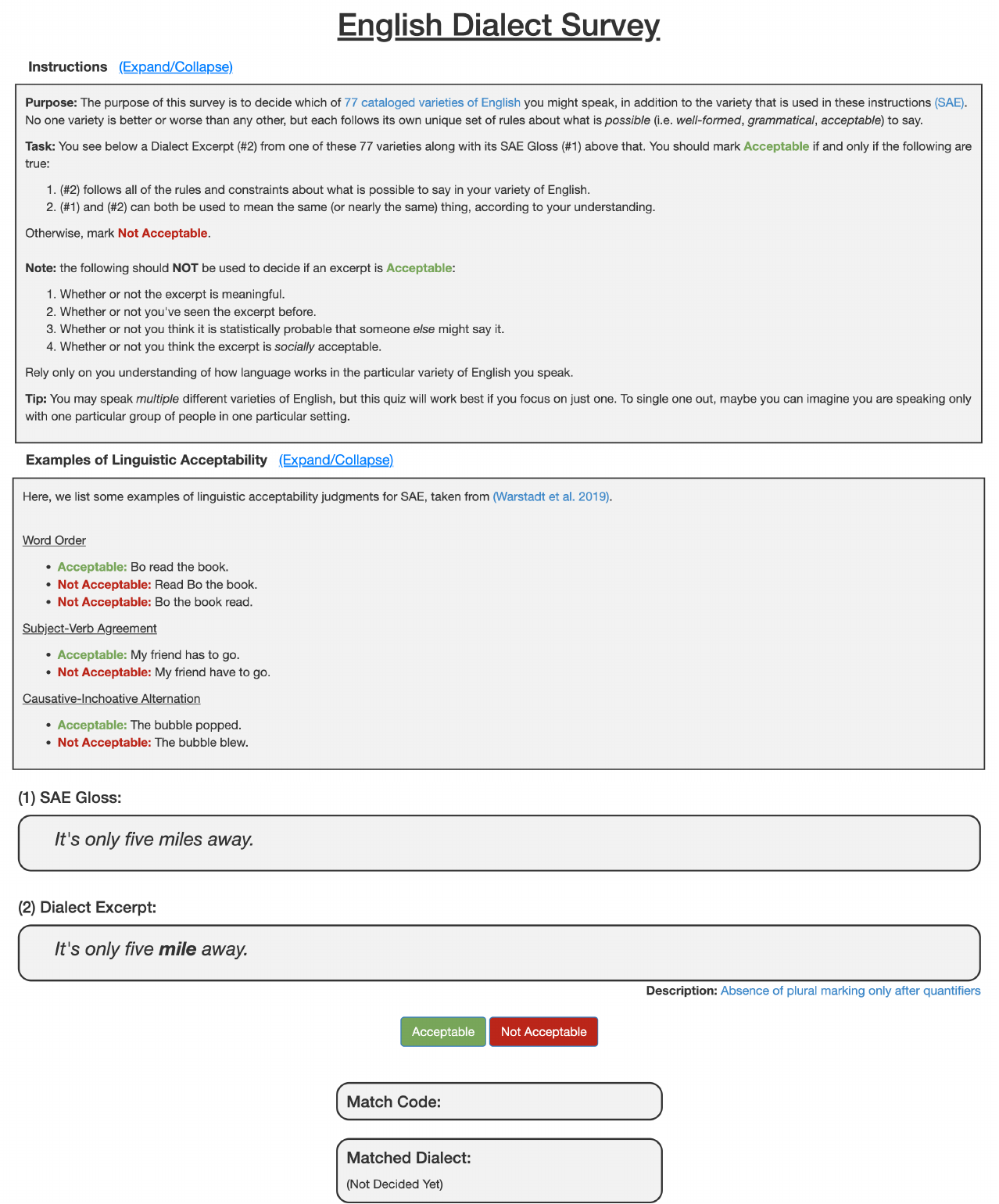}
    \caption{
    \textbf{The English Dialect Speaker Assessment Quiz} used for matching dialect speaker annotators to specific dialects for prompt annotation. We adapt the assessment quiz from the existing English Dialect Speaker Survey first created in MultiVALUE~\citep{Ziems2023MultiVALUEAF}, which asks the human annotator to select their linguistic acceptability preference for 10 different dialect excerpts.
    }
    \label{fig:assessment}
\end{figure*}

In the creation of the \Dataset{} Dataset, we recruit a total of 17 dialect speaker human annotators from Amazon Mechanical Turk. The demographic involves six annotators from Asia, eight annotators from North America, and 3 annotators from Europe. Each selected annotator is given the option to complete any number of questions as they prefer. We encourage each annotator to take regular breaks during the task and not to work consecutively for more than 2 hours on our task. Our task is relatively simple for dialect speakers as it mainly involves judging the plausibility and meaning of a sentence in their native dialect. We estimate each HIT to take around 12 seconds, this corresponds to an hourly wage of \$15 USD. Our total annotation time is 21.84 hours, costing a total of \$327.6. We ran 4 rounds of annotations, with a combined total of 6552 prompts. 35.9\% of total proposed prompts were rejected by the annotators while 64.1\% of prompts were approved.

\section{Use of AI tools}
We employed large language models (LLMs), including OpenAI’s GPT 5, as auxiliary tools to refine the manuscript and identify grammatical errors. All LLM-assisted content was critically reviewed, fact-checked, and revised by the authors to ensure scientific validity and originality. The authors retain full responsibility for all statements and conclusions presented in this work. Specifically, LLMs were used only to improve wording and clarity of expression.

\section{Limitations}
\label{sec:limitation}
Our study focuses on the lexical variations that characterize dialects, motivated by the empirical observation that such variations exert much greater influence on multimodal generative model performance than grammatical variations (see \Cref{subsec:grammatical}). Furthermore, grammatical variation has already been the subject of extensive investigation in text-only contexts~\citep{hudson1996sociolinguistics, chambers1998dialectology, fromkin1998introduction, nerbonne2009data, wardhaugh2021introduction}. These considerations jointly motivate our decision to prioritize the evaluation of lexical dialect variation, which appears especially consequential in the multimodal generative setting.
Furthermore, our evaluation of text-image alignment utilizes reference-free metrics, namely VQAScore~\citep{lin2024evaluating} and CLIPScore~\citep{Hessel2021CLIPScoreAR}. We recognize that these pretrained vision-language models are not perfect. To address this potential weakness, we conducted a thorough human evaluation and found very high statistical correlation between our automatic metrics and human judgment (Pearson correlation coefficient $r$ = 0.968 for VQAScore and $r$ = 0.924 for CLIPScore). Therefore, while acknowledging the imperfections of automated metrics, this high degree of human correlation provides strong evidence for the validity of our evaluation metrics and associated analysis conclusions.

\section{Future Work}
Our work highlights several promising directions for future research, which we encourage the community to explore.

\paragraph{Investigating Cultural and Representational Biases}
It would be interesting for future works to explore and evaluate the significance of representational and skin tone shifts induced by dialect inputs. For instance, as noted in \Cref{fig:teaser}, we observed that FLUX.1 [dev]~\citep{flux2024} image generations for the prompt ``A man selling eggplant'' depict more upscale and decorated environments compared to generations for ``A man selling brinjal.'' Furthermore, individuals depicted in the images for ``brinjal'' are darker-skinned. A systematic study of these shifts would provide valuable insights into the inherent biases of large-scale multimodal models.

\paragraph{Exploring Grammatical and Joint Dialect Variations}
While this work concentrated on lexical variations, we welcome future works in this line to carefully study the impacts of grammatical dialect variations and their joint effects with lexical variations. Such research could reveal more complex interactions and failure modes in the performance of multimodal generative models.

\paragraph{Investigating Downstream Impacts of Dialectal Performance Gaps}
Many existing research studies~\cite{zhang2023adding, wallace2024diffusion, geng2024unmet, zhou2025contrastive, li2025real} and industrial systems rely on the accurate semantic understanding and high-fidelity generation capabilities of multimodal text-to-image and text-to-video generative models. It would be interesting to investigate the downstream research impacts of dialectal performance gaps on these works as well as downstream societal impacts to dialect speaker user groups.

\paragraph{Extending Evaluation to Multi-Lexeme Prompts}
Another related area for future work is the extension of our evaluation to settings where multiple dialect lexemes are used. This would test the models' compositional understanding of dialectal language, and we encourage future works to explore such possibilities. However, it should be noted that creating high-quality, controlled data at scale for such experiments is a non-trivial problem that needs to be addressed.

\section{Societal Impact}
\label{sec:impact}

This work makes use of human subjects for annotation and evaluation. All procedures were subject to ethical review and were approved by the IRB from the authors’ institution. Consent was gathered in accordance with the authors’ institution guidelines, and annotators had access to a data use statement when giving consent. The purpose of \Dataset{} is to provide tools that enable researchers and practitioners to evaluate and improve dialect robustness in their models. We will release these data responsibly, ensuring that users sign a Data Use Agreement that forbids the use of \Dataset{} for deception, impersonation, mockery, discrimination, hate speech, targeted harassment, and cultural appropriation. In the agreement, researchers and practitioners will also acknowledge the limitations of this work, that \Dataset{} may not fully or accurately represent the natural usage patterns of all sub-communities of speakers. \Dataset{} is designed to be easily updatable and configurable, such that it can be extended by and for specific sub-communities and updated as dialects evolve over time. We have carefully checked our data to make sure no personally identifying information or offensive content is included. When utilizing existing artifacts and models, we make sure to follow all relevant regulations and licenses.

\section{Reproducibility}
\label{sec:reproducibility}

We have taken several steps to ensure the reproducibility of our work. Detailed descriptions of dataset construction, annotation procedures, evaluation protocols, and mitigation methods are provided in the main paper (see \Cref{sec:dataset}, \Cref{sec:experiments}, etc.), with further implementation details, training configurations, and additional qualitative results included in the appendix (see \Cref{subsec:qualitative_comparison}, \Cref{subsec:implementation_details}, etc.). To facilitate independent verification, we also provide as anonymized supplementary material both the DialectGen benchmark dataset and the source code used for data processing, model training, and evaluation. The dataset files include all validated dialect--SAE prompt pairs, while the code folder contains scripts for dataset generation, automatic and human evaluation, and reproduction of all tables and figures reported in the paper. Together, these resources enable researchers to replicate our experimental results and extend the benchmark for future work.

\end{document}